\documentclass[10pt,twocolumn,letterpaper]{article}
\usepackage[T1]{fontenc}    % use 8-bit T1 fonts
\usepackage{amsmath}
\usepackage{hyperref}       % hyperlinks
\usepackage{url}            % simple URL typesetting
\usepackage{amsmath,amssymb}
\usepackage{graphicx}
\usepackage{tikz,graphics,color,caption,subcaption}
\usepackage{tikz,graphics}
\usepackage{cvpr}

\begin{document}

%% TITLE
% \title{SqueezeSAM: From interactive segmentation to saliency and back} % inspired by this paper title From Captions to Visual Concepts and Back - arxiv.org/abs/1411.4952
\title{SqueezeSAM: User-Friendly Mobile Interactive Segmentation}
\author{Balakrishnan Varadarajan \and Bilge Soran  \and Forrest Iandola \and Xiaoyu Xiang \and  Chenchen Zhu \and Yunyang Xiong \and Lemeng Wu \and Raghuraman Krishnamoorthi \and Vikas Chandra \and {\tt\small \{balakv,bsoran,fni, xiangxiaoyu,yunyang,chenchenz,lmwu,raghuraman,vchandra\}@meta.com}}

\maketitle

%%%%%%%%% ABSTRACT
\begin{abstract}

The Segment Anything Model (SAM) has been a cornerstone in the field of interactive segmentation, propelling significant progress in generative AI, computational photography, and medical imaging. Despite its ability to process arbitrary user input and generate corresponding segmentation masks, SAM's 600 million parameter architecture, based on ViT-H, is not compatible with current mobile hardware due to its high computational demands and large model size. Our research aims to adapt SAM for use in mobile photography applications. To this end, we have developed a fully convolutional SqueezeSAM model architecture, which is 62.5 times faster and 31.6 times smaller than the original SAM, making it a viable solution for mobile applications. Furthermore, our tiny model achieves an mIOU within \emph{1\%} of the original VIT-H architecture.

Automated segmentation holds significant value in the creation flow for photography applications, as evidenced by its adoption by leading industry players like apple and capcut. To facilitate this automation, we employ salient object detection and simulate potential user clicks for foreground object selection, generating an initial segmentation mask that users can subsequently edit interactively. A common user expectation is that a click on a specific part of an object will result in the segmentation of the entire object. For example, a click on a person's t-shirt in a photo should ideally segment the entire person, not just the t-shirt. However, SAM typically only segments the clicked area. We address this limitation through a novel data augmentation scheme. Consequently, if a user clicks on a person holding a basketball, both the person and the basketball are segmented together, aligning with user expectations and enhancing the overall user experience.

\end{abstract}
\section{Introduction}
\label{sec:intro}
Since the introduction of the Segment Anything Model (SAM)~\cite{Kirillov2023_SAM}, it has catalyzed significant advancements in generative AI, medical imaging, and computational photography~\cite{shen2023_anything3d, ma2023_MedSAM, yu2023_inpaint}. Our research focuses on interactive segmentation, a feature that enables users to select and extract objects within images on a smartphone. This functionality has already been implemented in the iPhone with iOS 16, but the proprietary model used by Apple is not available for retraining by researchers. Our objective is to develop an open-source, fast, user-friendly, mobile model for interactive image segmentation. 

The adoption of automated segmentation by major industry players underscores its substantial importance for photography applications like apple and capcut. SAM requires the user to initiate the segmentation process by clicking on an object. To overcome this, we propose using salient object detection (SOD) to select a few interesting points in the image, enabling SAM to generate an initial segmentation that can be refined through user input. As SOD aims to identify objects and regions that attract human attention, it is logical to use it to predict the image regions that humans are likely to click on.

Another challenge with SAM is its latency -it takes half a second on an A100 GPU and over ten seconds on an iPhone CPU. Local device operation is crucial for interactive latency and privacy preservation. The latency of SAM is primarily due to its encoder, a ViT-H vision transformer~\cite{dosovitskiy2021_ViT}. While more efficient SAM variants exist, they significantly compromise the quality-of-results~\cite{cai2023_efficientvit, zhang2023_MobileSAM, zhao2023_FastSAM, NanoSAM}. In our work, we explore model architectures and training schemes that offer low latency while maintaining a quality-of-results competitive with the original SAM model. We also show that our models deliver higher quality segmentation of salient objects.

In summary, our novel contributions include a new model architecture that is \emph{62.5 times faster and 31.6 times smaller} than the original SAM, and the ability to generate segmentation masks for salient objects without user inputs. We will also publicly release the weights of our compact model, trained on 1 billion masks and 11 million images.

\begin{figure}
  \centering
  \includegraphics[width=80mm]{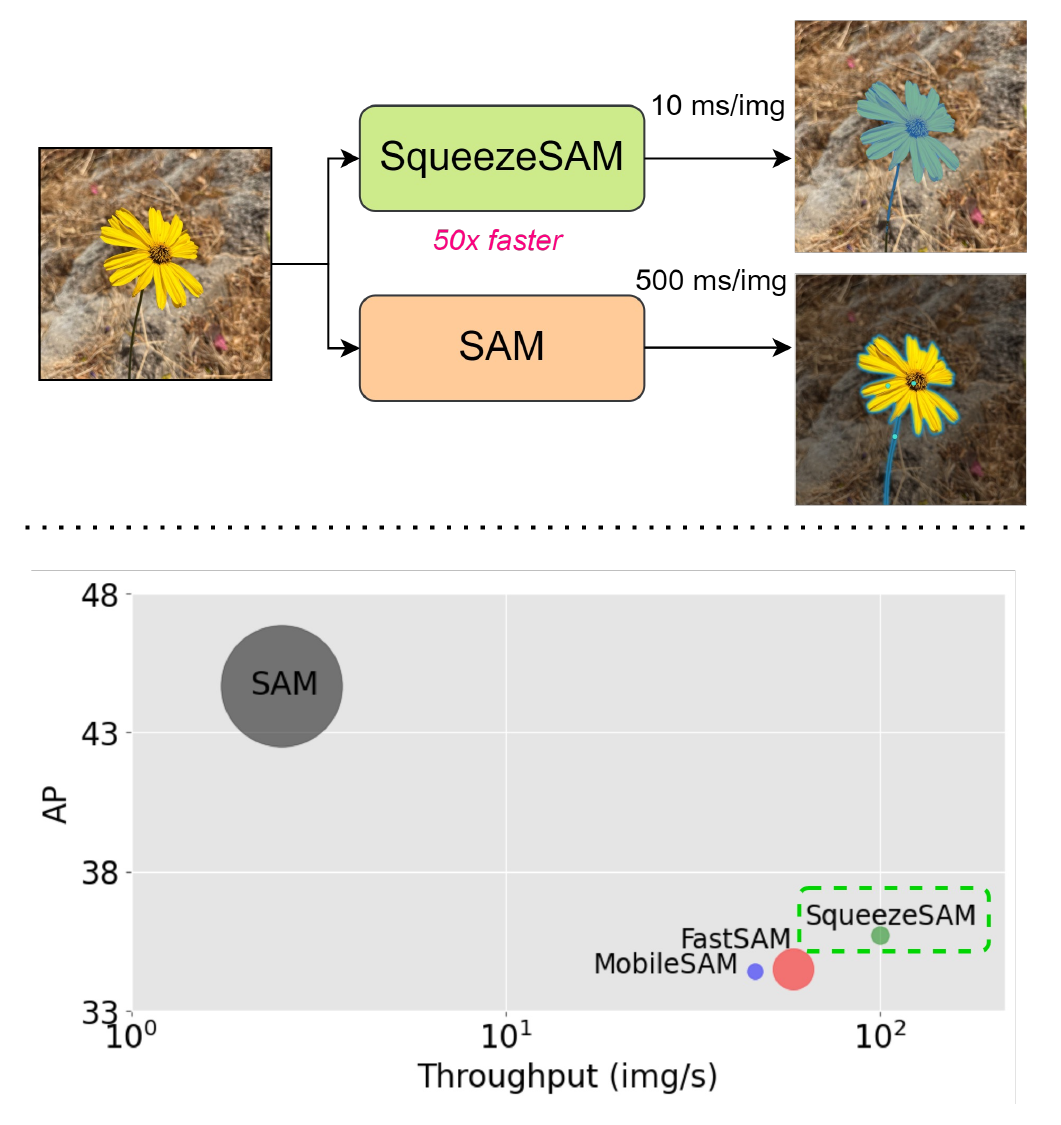}
  \caption{A comparative analysis of SqueezeSAM and SAM. (Top) Speed comparison between SqueezeSAM and SAM on a single NVIDIA A100 GPU. (Bottom) Performance/Runtime/Parameter comparison of SAM and other efficient models.}
  % \caption{The comparative analysis result. (Top) The overview of EfficientSAM model by taking a well-pretrained light-weight image encoder for instance segmentation with largely reduced complexity. (Bottom) Speed/Parameter/Performance comparison of EfficientSAM, MobileSAM, FastSAM, and SAM for zero-shot instance segmentation on COCO. We benchmark throughput (image per second) of all models on a single NVIDIA A100 with one box prompt. Our EfficientSAMs outperform MobileSAM and FastSAM by a large margin, $\sim $4 AP, with comparable complexity. Our EfficientSAM-S reduces the inference time of SAM by $\sim $10x and the parameter size by $\sim $20x with a small performance drop, 44.4 AP vs 46.5 AP. }
  \label{fig:throughput}
\end{figure}

\begin{figure*}
    \centering
    % figure source: https://app.diagrams.net/#G1IK-IrExYT6XpU2dyw1oewavR2YNEBCqf
    % \fbox{
        \includegraphics[width=\linewidth]{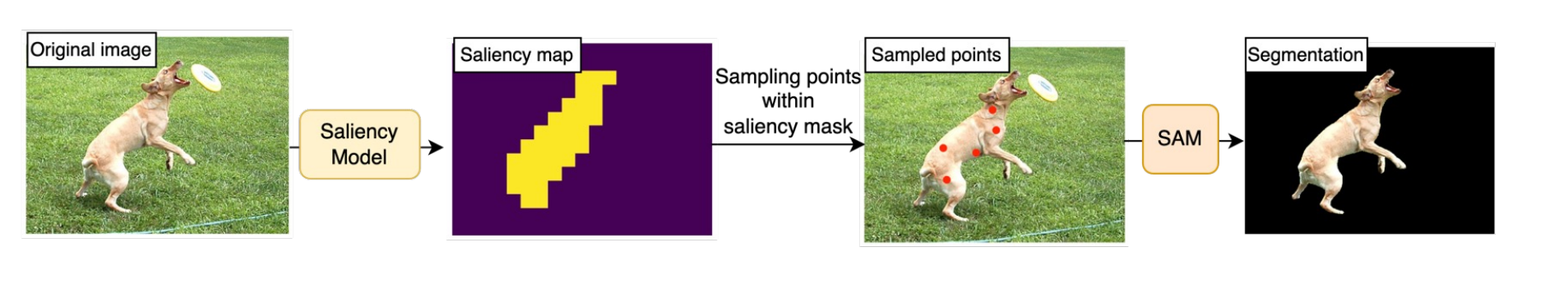}
    % }
    \caption{\textbf{Salient SqueezeSAM pipeline}. Salient Object Detection feeds initial points to SqueezeSAM, so the user can get an initial segmentation without clicking on the screen. After that, the user can interactively edit and improve the segmentation. It runs in real-time on an iPhone.}
    \label{fig:pipeline}
\end{figure*}

\section{Related work}
{\bf Interactive segmentation} is a task where a user clicks on interesting regions of an image, and a computer vision model segments those regions.
From the 1980s through the early 2010s, Snakes~\cite{Kass1988_Snakes}, GraphCut~\cite{Boykov2001_GraphCut}, and Geodesic Star~\cite{Gulshan2010_geodesic_star} applied classical computer vision techniques to this problem.
Then, deep learning spread from object classification~\cite{Krizhevsky2012_AlexNet} to semantic segmentation~\cite{long2015_FCN} to interactive segmentation~\cite{xu2016_DIOS}.

By 2022, the de facto approach to interactive segmentation was the following.
The model architecture consisted of a Vision Transformer (ViT)~\cite{vaswani2017_transformer, dosovitskiy2021_ViT} or ConvNet encoder plus a feature pyramid network decoder~\cite{lin2016_FPN}.
The training set incorporated COCO~\cite{lin2015_coco}, LVIS~\cite{gupta2019_LVIS}, and possibly a few other instance segmentation datasets.
Evaluation datasets included LVIS and DAVIS~\cite{Zeng_2019_davis_s}.
To evaluate the model, a standardized algorithm simulates mouse-clicks on an image, and it stops clicking when the generated segmentation masks achieve an intersection over union (IOU) of 0.9.
The fewer clicks to achieve 0.9 IOU, the better.
% An other way to evaluate was to report the mean IOU that is achieved with a certain number of clicks.
In this regime, the leading models included FocalClick~\cite{chen2022_focalclick}, SimpleClick~\cite{liu2023_simpleclick}, and RITM~\cite{sofiiuk2021_RTIM}.
To segment images with 0.9 IOU, SOTA deep neural networks require 5 times fewer clicks than classical approaches~\cite{sofiiuk2021_RTIM}.

\noindent {\bf Segment Anything Model (SAM)} has significantly advanced the field of interactive segmentation~\cite{Kirillov2023_SAM}. Like SimpleClick, SAM employs a ViT encoder. However, in contrast to SimpleClick's feature pyramid network decoder, SAM utilizes a Transformer-based decoder. The most notable departure of SAM from previous work is its training on a novel dataset, SA-1B, comprising 11 million images and 1 billion masks. This new dataset was instrumental in enabling SAM to surpass FocalClick, SimpleClick, and RITM in both interactive and instance segmentation benchmarks.

Much work has been built on top of SAM.
% HQ-SAM adds a side-network that refines and improves SAM predictions~\cite{ke2023_SAM_HQ}.
SAM is core to generative AI tasks including style transfer~\cite{Psychogyios2023_SAMStyler, liu2023_anytoany}, image inpainting~\cite{yu2023_inpaint}, text guided image and video editing~\cite{xie2023_EditEverything, wu2023_tgve}, and 3D reconstruction and generation~\cite{shen2023_anything3d, shi2023_zero123plusplus}.
Further, SAM has been applied to general medical image segmentation~\cite{ma2023_MedSAM} and specialized tasks like polyp segmentation~\cite{zhou2023_Polyps}, skin lesion segmentation~\cite{wu2023selfprompting}, and COVID-19 lung imagery~\cite{cheng2023_sam_medical}.
Improvements to the quality and efficiency of SAM will benefit all of these applications.

\noindent {\bf Alternative SAM model architectures} have been proposed for faster inference.
While SAM uses a large vision transformer called ViT-H, MobileSAM uses the ViT-tiny encoder model~\cite{zhang2023_MobileSAM}.
Others use convolutional encoder models: FastSAM uses a variant of the YOLOv8 encoder architecture~\cite{Jocher2023_YOLOv8, zhao2023_FastSAM}, and NanoSAM uses ResNet18~\cite{he2015_resnet, NanoSAM}.
EfficientViT uses an encoder that includes transformers and convolutions~\cite{cai2023_efficientvit}.
To avoid the high cost of training from scratch on the SA-1B dataset, MobileSAM, NanoSAM, and EfficientViT train via distillation from SAM.
FastSAM trains on a small subset of the SA-1B dataset.

\noindent {\bf Salient Object Detection (SOD)} involves predicting a heat map that distinguishes between salient and non-salient image regions. \footnote{To dispel a common point of confusion: While object detection typically means bounding boxes, SOD uses masks.}
Since 2010, several SOD benchmark datasets have emerged.
In most benchmark datasets, humans label the ground-truth masks using a graphical interface~\cite{Movahedi2010_SOD_dataset, wang2017_duts_te, Qiong2013_ECSSD}.
Some datasets also incorporate human eye tracking data into the labeling process, based on the premise that areas of fixation are likely to be salient~\cite{li2014_pascal_s, yang2013_dut_omron}.
Top performing models on SOD benchmarks include R$^3$Net~\cite{Deng2018_r3net}, U$^2$Net~\cite{Qin_2020}, and SSOM~\cite{cui2023_SSOM}.
Notably, SSOM employs SAM to enhance salient object detection. In a related vein, our research utilizes SOD to augment interactive segmentation.

\section{Proposed approach}

The task of zero-shot instance segmentation using user clicks can be formally defined as follows. Given an image $\mathbf{I} \in \mathbb{R}^{3 \times H \times W}$ and a set of user clicks $\mathbf{u} = (x_c,y_c)_{c=1,|C|}$, the segmentation problem is to provide the probability of every pixel in the image belonging to the mask as $p(b_{ij} | (x_c,y_c)_{c=1,|C|})$.

The original SAM model consists of an encoder $\mathcal{E}$ and a decoder $\mathcal{D}$.
Given an image $\mathbf{I}$, the encoder first embeds the image into an rich intermediate representation $\mathbf{x} = \mathcal{E}(\mathbf{I})$.
The decoder (typically light weight) consumes the intermediate representation $\mathbf{x}$ and an external user input $\mathbf{u}$ to produce a set of $k$ segmentation mask $\mathbf{m} \in \mathbb{R}^{H \times W \times k}$ along with their estimated IOU scores $\mathbf{s} \in \mathbb{R}^{k}$. The proposed idea of early fusion combines the input to the encoder with user clicks as $\mathbf{I} \times \mathbf{u}$. We will summarize the motivations for early fusion in the next few sections.

\begin{figure*}[!bhpt]
    \centering
    \includegraphics[width=\textwidth]{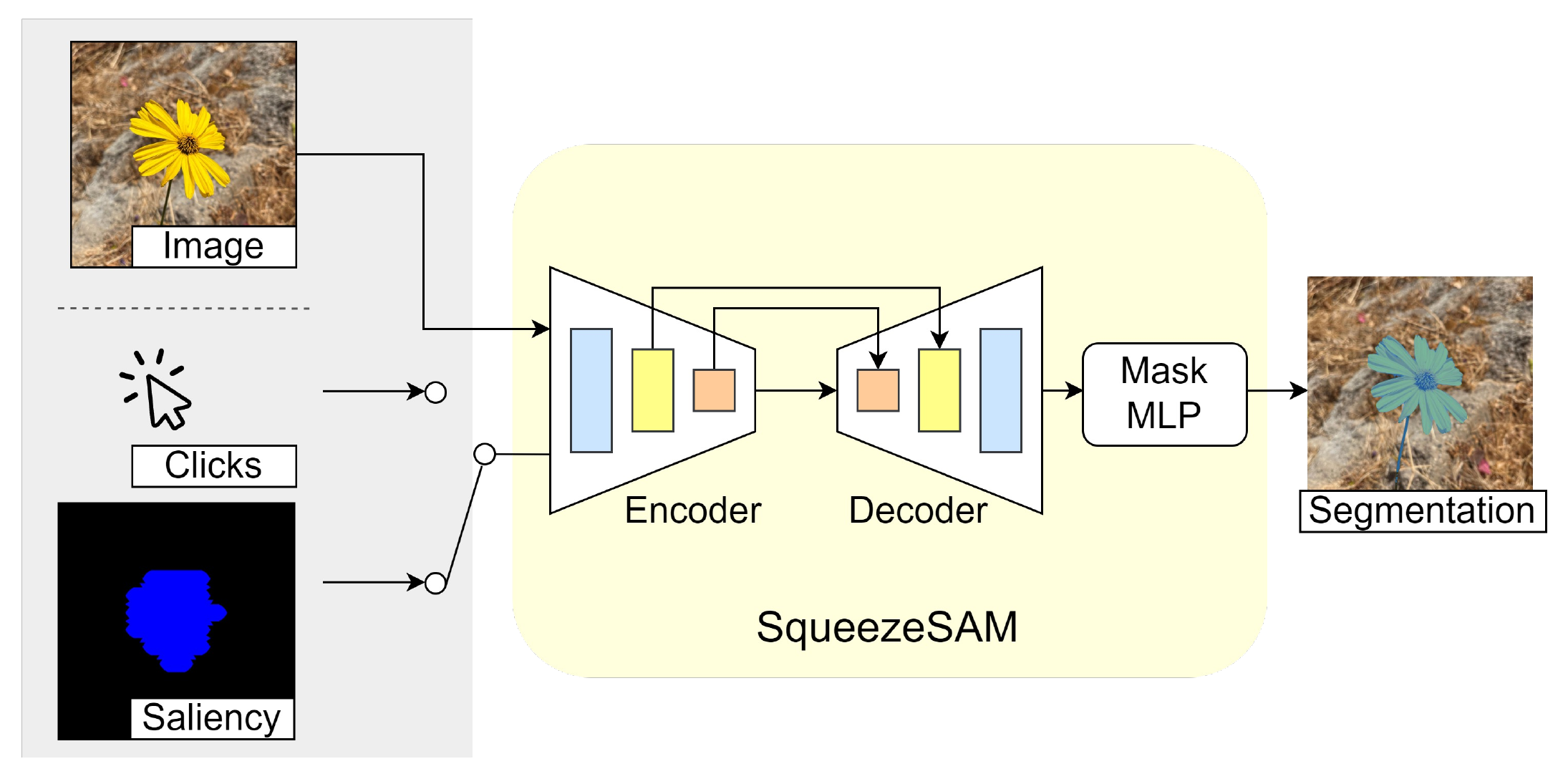}
    \caption{SqueezeSAM model architecture. Our proposed SqueezeSAM can take point coordinates from user's clicks, or ``guess" user's intention from a saliency mask, and generate the corresponding segmentation output. For example, the model can begin by segmenting one object, and the user can interactively edit the segmentation or segment more objects.}
    \label{fig:E2E_Architecture}
\end{figure*}

\begin{figure*}
    \centering
    % edit this diagram here: https://app.diagrams.net/#G1jtwuGQgkEqpzutsa6NQyG5uzl4Ww9y_T
    % \fbox{
        \includegraphics[width=12cm]{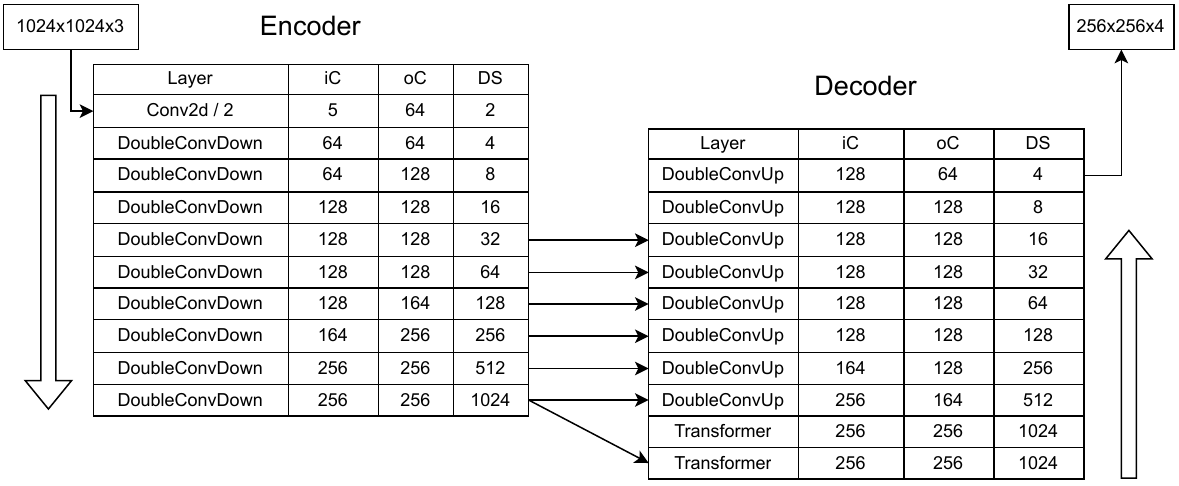}
    % }
    \caption{SqueezeSAM encoder-decoder model architecture. DoubleConv modules are described in later figures. iC = input channels from the previous layer, not including concatenated skip-connections. oC = output channels. DS = downsampling factor, relative to the original input image.}
    \label{fig:SqueezeSAM_dims}
\end{figure*}

\begin{figure}
    \centering
    % edit this diagram here: https://app.diagrams.net/#G1jtwuGQgkEqpzutsa6NQyG5uzl4Ww9y_T
    \fbox{
        \includegraphics[width=5cm]{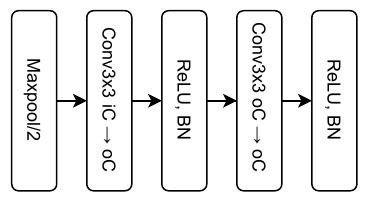}
    }
    \caption{{\bf DoubleConvDown} module used in the SqueezeUNet encoder.}
    \label{fig:DoubleConvDown}
\end{figure}

\begin{figure}
    \centering
    % edit this diagram here: https://app.diagrams.net/#G1jtwuGQgkEqpzutsa6NQyG5uzl4Ww9y_T
    \fbox{
        \includegraphics[width=6cm]{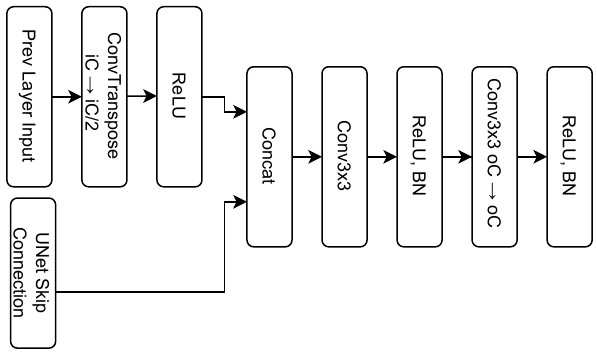}
    }
    \caption{{\bf DoubleConvUp} module used in the SqueezeSAM decoder. The ConvTranspose does 2x upsampling and reduces the channels by 2x.}
    \label{fig:DoubleConvUp}
\end{figure}

\subsection{SqueezeSAM Encoder-Decoder Architecture}
Our initial exploration began with using a VIT tiny backbone (as an alternative to VIT-H) trained on the SA-1B dataset. Due to the difficulty in quantization and deployment of VIT tiny models, subsequently, we also explored a UNet based architecture and discovered that we can train efficient and high quality models easily using a UNet backbone without much architecture tuning.  We show our end-to-end model architecture in Figure~\ref{fig:E2E_Architecture}.
The inputs to our model are an image $\mathcal{I} \in \mathbb{R}^{3 \times 1024 \times 1024}$ and a set of clicks.\footnote{For now, let's assume the clicks are from a user clicking a mouse or tapping a smartphone screen. In Section~\ref{sec:salient_squeezesam}, we will explain how to use salient object detection to synthesize clicks.}
The model's computational footprint is dominated by the SqueezeSAM encoder and decoder.
In Figure~\ref{fig:SqueezeSAM_dims}, we show the precise layers and dimensions of our encoder and decoder, and we show the details of the DoubleConv blocks in Figures~\ref{fig:DoubleConvDown} and~\ref{fig:DoubleConvUp}.
Here are a few key design choices we made and why we made them:

\begin{itemize}
    \item {\bf Apply transformers at the bottom scale of the UNet.} We downsample the image from 1024$\times$1024$\times$3 all the way down to 1$\times$1$\times$256. We then apply two transformer layers on the 1$\times$1$\times$256 representation. This is informed by papers such as SnapFusion~\cite{2023_li_snapfusion} and NASViT~\cite{2022_gong_nasvit}, which find that when combining convolutions and transformers, the best speed-accuracy tradeoff is achieved by using transformers in the lower-resolution layers.
    \item {\bf Low channel count.} In addition to optimizing for latency, we wish to maintain a small model size to avoid wasting storage space on the smartphone. So, while some UNet models double the number of channels each time they downsample (going over 1000 channels)~\cite{2023_li_snapfusion}, we use a maximum of 256 output channels per layer.
    \item {\bf BatchNorm.} While works such as ConvNext~\cite{2022_liu_convnext} show the accuracy advantage of LayerNorm over BatchNorm, there are two computational-efficiency advantages of using BatchNorm. First, LayerNorm has a square-root operation, which we find is particularly expensive on mobile hardware (adding up to 50\% to the total latency). Second, after training is complete, BatchNorm can be folded into a Conv+Bias layer, so there is no computational cost to BatchNorm inference~\cite{2018_krishnamoorthi_quantizing}. This folding is not possible with LayerNorm.
    \item {\bf Skip connections.} Following prior literature on UNet models~\cite{2015_ronneberger_unet}, we add skip connections from the hidden layers of the encoder to the hidden layers of the decoder.
\end{itemize}

After the encoder and decoder, we adopt the Mask MLP block from original SAM. It has one small MLP for each of the 4 output channels.

% Use of batch normalization at various places:
% We find that using normalization at various places in the net improves convergence and quality. We also found that using batch norm is able to achieve much better latency than layer norm (about 50\%).

\subsection{Early fusion}
We add the user input points as separate channels (along with the original RGB image input). In particular, these clicks are encoded as circles in the fourth channel and the bounding box is encoded as a rectangle in the fifth channel. By knowing the clicks upfront, the model is able to focus more of its attention and its parameters on the image regions that user wants to segment. In addition to this early fusion, points are also encoded using the decoder transformer (as is done with SAM). The mix of early and late fusion helps the net further in being able to attend to the sought objects.

\subsection{Training}
We train our models on the SA-1B dataset for 10 epochs using V100 GPUs. We use a batch size of 4 per GPU (effective batch size of 256) and synchronized batch norm applied between each of the UNET layers. During training, we sample 8 masks per image uniformly at random.  An initial learning rate of $5\mathrm{e}{-4}$ is followed by a linear decay is employed. The total training time takes around 7 days.
%Once the model finishes training on SA-1B, we finetune the model on LVIS/COCO images. --> We did finetune some models on LVIS/COCO, but not the ones for this section of the paper

% \bala{It sounds like you want to claim an improvement on training efficiency. Can we clearly state the following? With original SAM, how many training steps, at what batch size, on how many GPUs, takes how many days?  With SqueezeSAM (early fusion), we say later in the paper that we train on SA-1B for 10 epochs, 64 GPUs, batch size of 8 per GPU ... now, how long does this take to train? Also, do we use fp16? How about torch.compile?  - Answer: Training for 5 epochs takes around 5 days. But this is dominated by data IO. We did not try fp16 and torch.compile. They will give some speedup, but not an order of magnitude due to data bottleneck. Note that GPU inference latency is around 20-30ms which is quite small. So forward/backward steps are actually not a bottleneck for our models.}

To achieve this relatively short training time on a large dataset, we changed how we feed clicks into the model during training.
In the original SAM training, for each batch, the model runs on one click, then another click point is sampled and the model runs again.
So, original SAM runs multiple steps of training for each batch of data.
This inflates the training time, so for SqueezeSAM we simply input a collection of points for each batch and run one training step for each batch. Otherwise, we adhere to the original training protocol from the SAM paper.

% We took the training recipe of the original SAM, identified bottlenecks and rectified. In the process, we sacrificed some of the original SAM functionalities (such as having the ability to take mask from previous iteration, iterative training/inference, etc). But we find that the gains obtained by improved training speed  are much higher since it facilitates using larger batch sizes, extensive model explorations, etc.
% \bala{When we say that we ``sacrificed some of the original SAM functionalities," is Late Fusion the only thing we sacrificed? Or is there something else? - I would not include late fusion as sacrifice. But (a) Intaking mask from previous iteration, (b) Use negative points}

% Some example key optimizations that were made are listed below:

% Batching: As an example we batched the decoder inference resulting in better GPU utilization. Batching of the decoder inference is not present in the original SAM implementation and a fair amount of model surgery was required to be able to make it batchable.
% Using an ideal number of dataset workers to balance GPU speed with data loading speed.
% The original SAM model accumulates loss across various examples using a mix of batching and for loops. We batched these loss computations to make it more GPU efficient.
% Other mundane optimizations such as using optimal jpeg compressions for the training images (in SA-1B), minimizing the number of opencv conversions, caching, etc also resulted in non-trivial improvements in training speed. % removed: using a manifold bucket that supports higher QPS,

\section{Evaluation}
To evaluate SqueezeSAM's interactive segmentation capability, we choose two instance segmentation benchmarks: COCO and LVIS.
Following the evaluation method of SAM~\cite{Kirillov2023_SAM}, we take bounding boxes generated from ViTDet~\cite{2022_li_vitdet} object detection, and we feed these bounding box into SqueezeSAM.
We use the same protocol when evaluating models from the literature.

To evaluate inference latency, we select the A100 GPU and the iPhone 14 CPU.
Why not evaluate on iPhone GPU or NPU?
There are thousands of models of smartphones, hundreds of smartphone brands, and dozens of manufacturers of smartphone chips.
As described in~\cite{2019_wu_ai_edge}, mobile GPUs and accelerators are not standardized, so when when deploying a model to all types of smartphones it's easier to use the CPU.

\begin{table*}[]
\caption{Instance segmentation results, latency and number of parameters. All compared models are prompted with ViTDet boxes to do zero-shot segmentation. AP numbers are reported as all, S, M, L. AP$^\text{P}$ denotes AP for the person category. AP and mIOU evaluation of FastSAM and EdgeSAM are from~\cite{Kirillov2023_SAM, zhao2023_FastSAM}, while the evaluation of SAM ViT-H, MobileSAM and EfficientSAM was done by the authors of this paper.}

\label{T:eval_standard_benchmarks}
\centering
\resizebox{\textwidth}{!}{%
% generated with https://www.tablesgenerator.com/
\begin{tabular}{l|cccccc|ccccc|cc|cc}
\hline
           & \multicolumn{6}{c|}{COCO}                & \multicolumn{5}{c|}{LVIS}             & \multicolumn{2}{c|}{latency (ms)} &   Params(M)       \\
     & mIOU & AP  & AP$^\text{L}$    & AP$^\text{M}$    & AP$^\text{S}$    & AP$^\text{P}$    & mIOU & AP  & AP$^\text{L}$     & AP$^\text{M}$     & AP$^\text{S}$     & A100           & iPhone          &  &  \\ \hline
SAM ViT-H~\cite{Kirillov2023_SAM}  &   78.4   & 46.5 & 61.7 & 51   & 30.8 & 53.5 &  78.9    & 44.7 & 65.5 & 57.6 & 32.5    & 500        &          -       &          600 \\ \hline
MobileSAM~\cite{zhang2023_MobileSAM}  &  74.3   & 38.7 & 54.3 & 42.2 & 23.7 & 39.0 &    72.3  & 34.4 & 53.7 & 44.9 & 23.8    &            12   &        -         &          9.7 \\
FastSAM~\cite{zhao2023_FastSAM}    &   -   & 37.9 & 50.0   & 43.4 & 23.9 &  -    &    -  & 34.5 & 50.0   & 43.4 & 23.9    &       -         &     -            &          68 \\
EdgeSAM~\cite{edgesam} &   76.7  & 43.0 & 55.1 & 48.9 & 30.3 & - & 76.0 & - & - & - & - & -  &  -  & 9.6    &  \\
EfficientSAM-Tiny~\cite{efficientsam} & 75.7 & 42.3 & 57.4 & 46.2 & 26.7 &  & 74.3 & 39.9 & 59.9 & 51 & 28.9 & 19  &  - & 10   &  \\
EfficientSAM-Small~\cite{efficientsam} & 77.0 & 44.4 & 60.1 & 48.3 & 28.4 &  & 75.4 & 42.3 & 62.3 & 54 & 30.8 & 21  &  - & 25   &  \\
SqueezeSAM (fp32) & 77.9 & 41.1 & 57.0 & 45.1 & 25.1 & 44.5 &  78.1   & 37.4 & 58.0 & 49.6 & 25.5 & 8  & 300 & 19 &  \\
SqueezeSAM (int8) & 77.5  & 40.8 & 56.7 & 45.0 & 24.6 & 43.4 &  77.8   & 36.9 & 57.6 & 49.2 & 24.6 & -  & 300 & 19 &  \\
\hline

\end{tabular}
}
\end{table*}

The disk size of our model is reduced to 14MB with int8 quantization. We find that with our architecture, we find less than 1 percent drop in model quality.

In Table~\ref{T:eval_standard_benchmarks}, we compare original SAM, SqueezeSAM, and MobileSAM.
We observe that, relative to the other efficient models, SqueezeSAM produces higher-quality results. Our model has a UNET architecture that is trained from scratch on the SA-1B dataset. Unlike most other techniques that use tricks like pre-training and distillation, our models can get competitive performance  without any of these techniques. We also find that our model has smaller latency compared to the MobileSAM version.

\newcommand{\qualiwidth}{0.16}
\begin{figure*}[htbp]
\captionsetup[subfigure]{labelformat=empty}
\begin{center}
  \begin{subfigure}[b]{\qualiwidth\linewidth}
  \includegraphics[width=\linewidth]{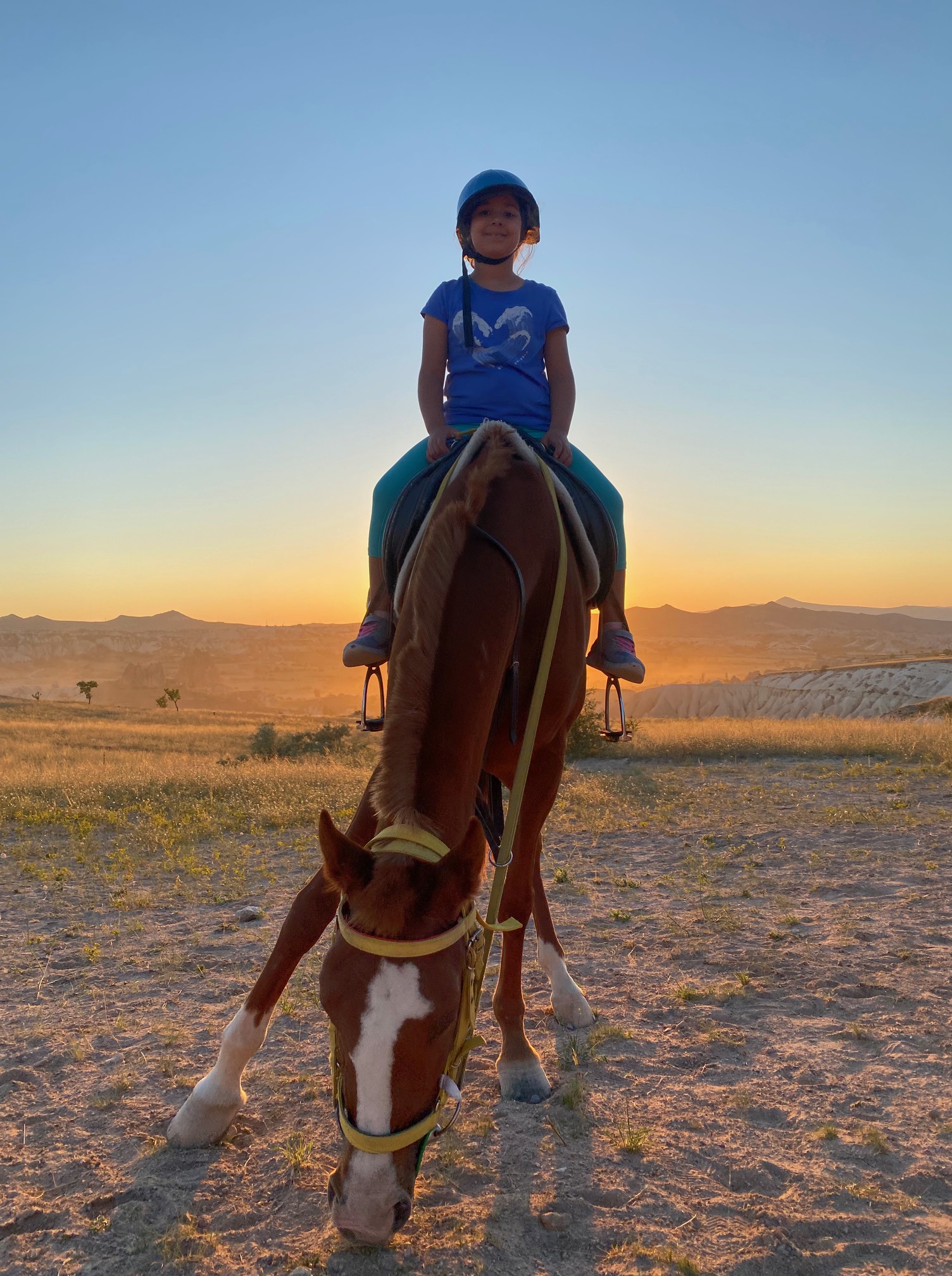}
  \end{subfigure}
\begin{subfigure}[b]{\qualiwidth\linewidth}
 \includegraphics[width=\linewidth]{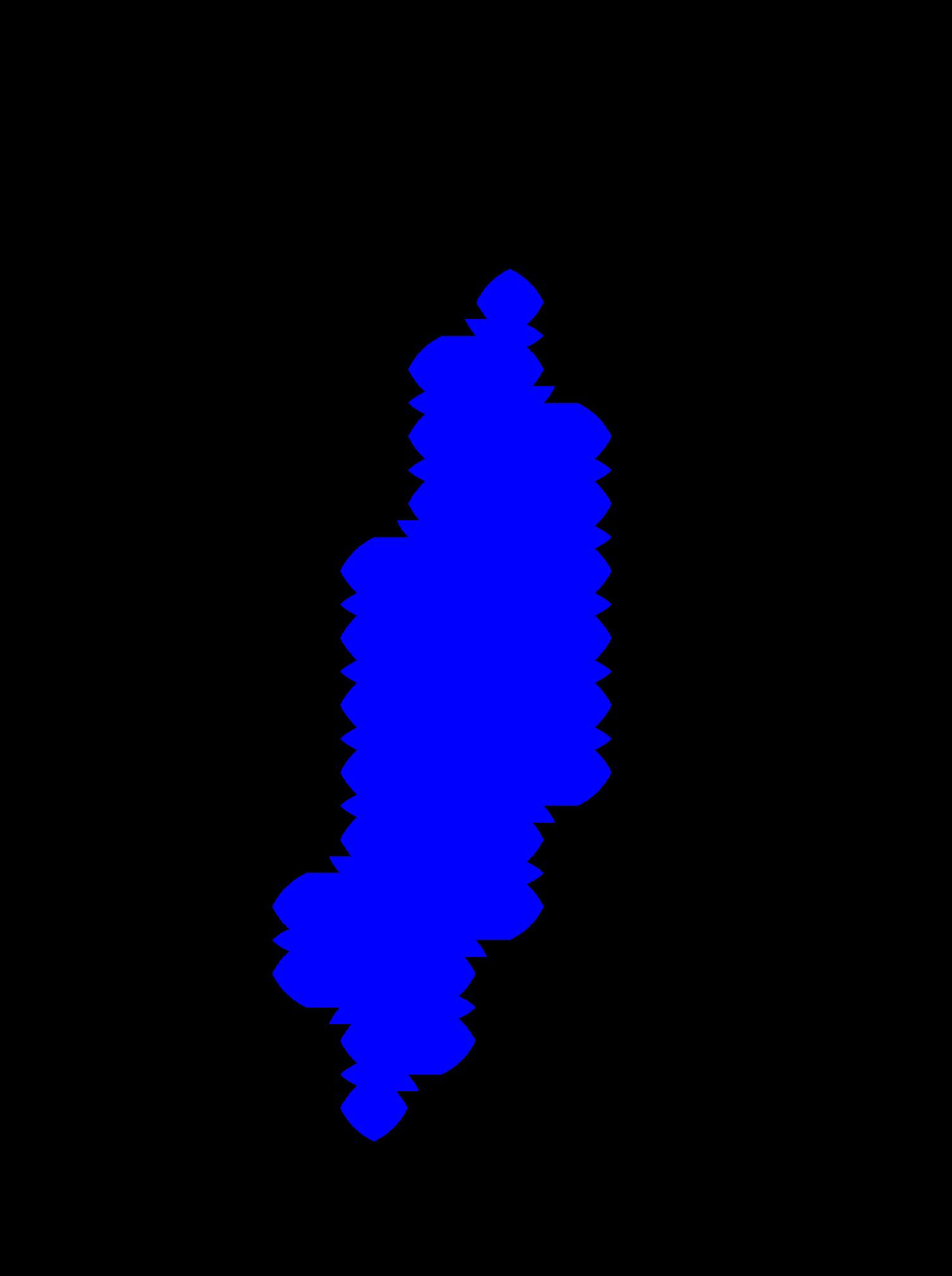}
  \end{subfigure}
  \begin{subfigure}[b]{\qualiwidth\linewidth}
  \includegraphics[width=\linewidth]{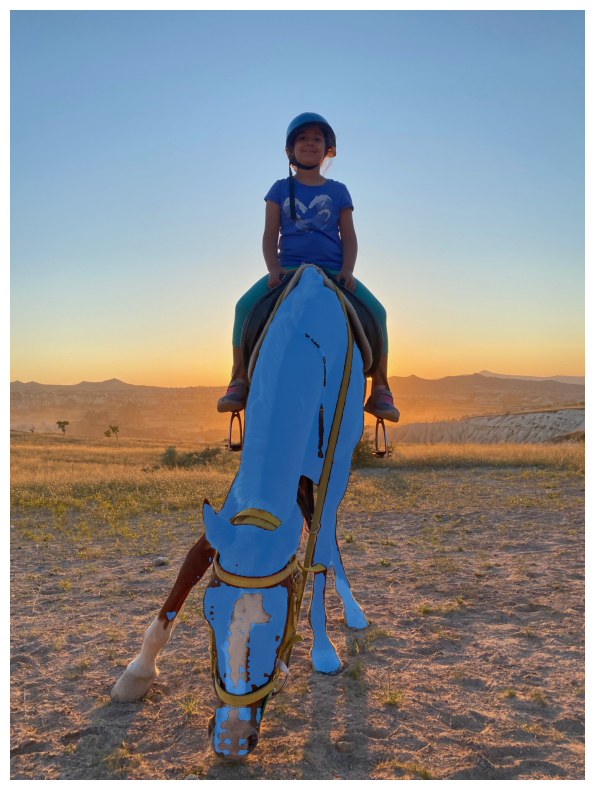}
  \end{subfigure}
  \begin{subfigure}[b]{\qualiwidth\linewidth}
 \includegraphics[width=\linewidth]{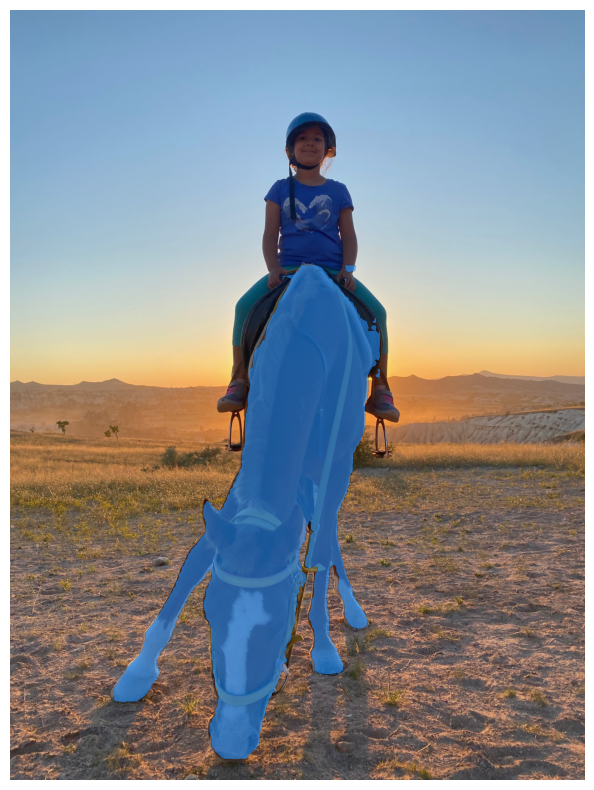}
  \end{subfigure}
  \begin{subfigure}[b]{\qualiwidth\linewidth}
  \includegraphics[width=\linewidth]{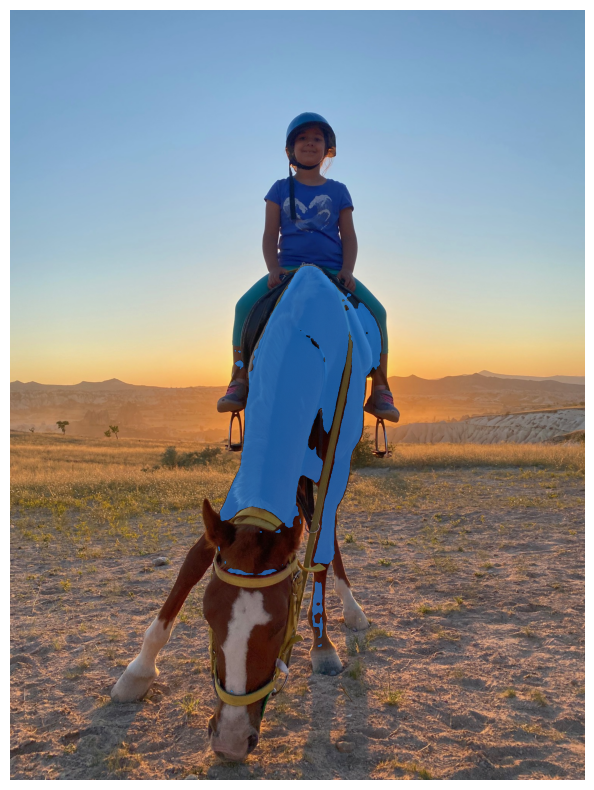}
  \end{subfigure}
  \begin{subfigure}[b]{\qualiwidth\linewidth}
 \includegraphics[width=\linewidth]{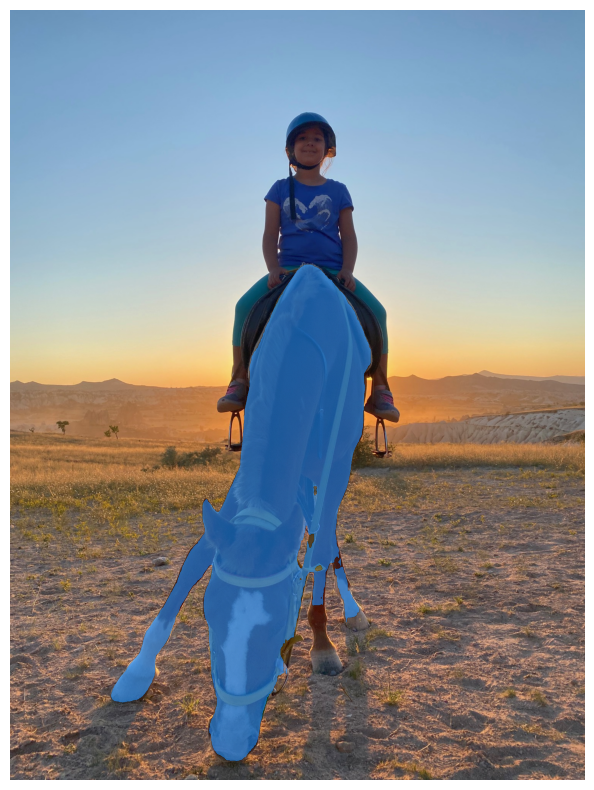}
  \end{subfigure}

\begin{subfigure}[b]{\qualiwidth\linewidth}
  \includegraphics[width=\linewidth]{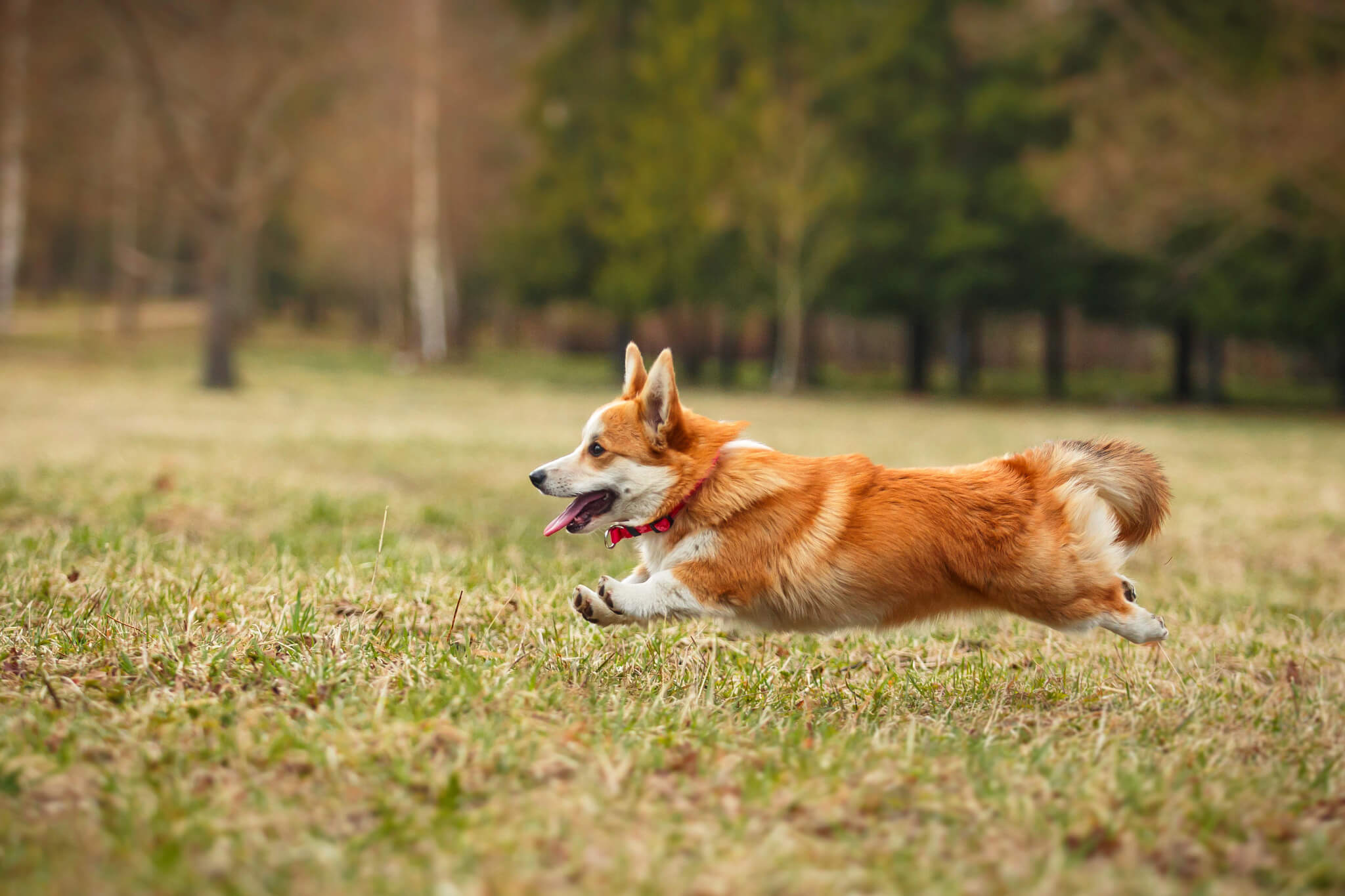}
  \end{subfigure}
\begin{subfigure}[b]{\qualiwidth\linewidth}
 \includegraphics[width=\linewidth]{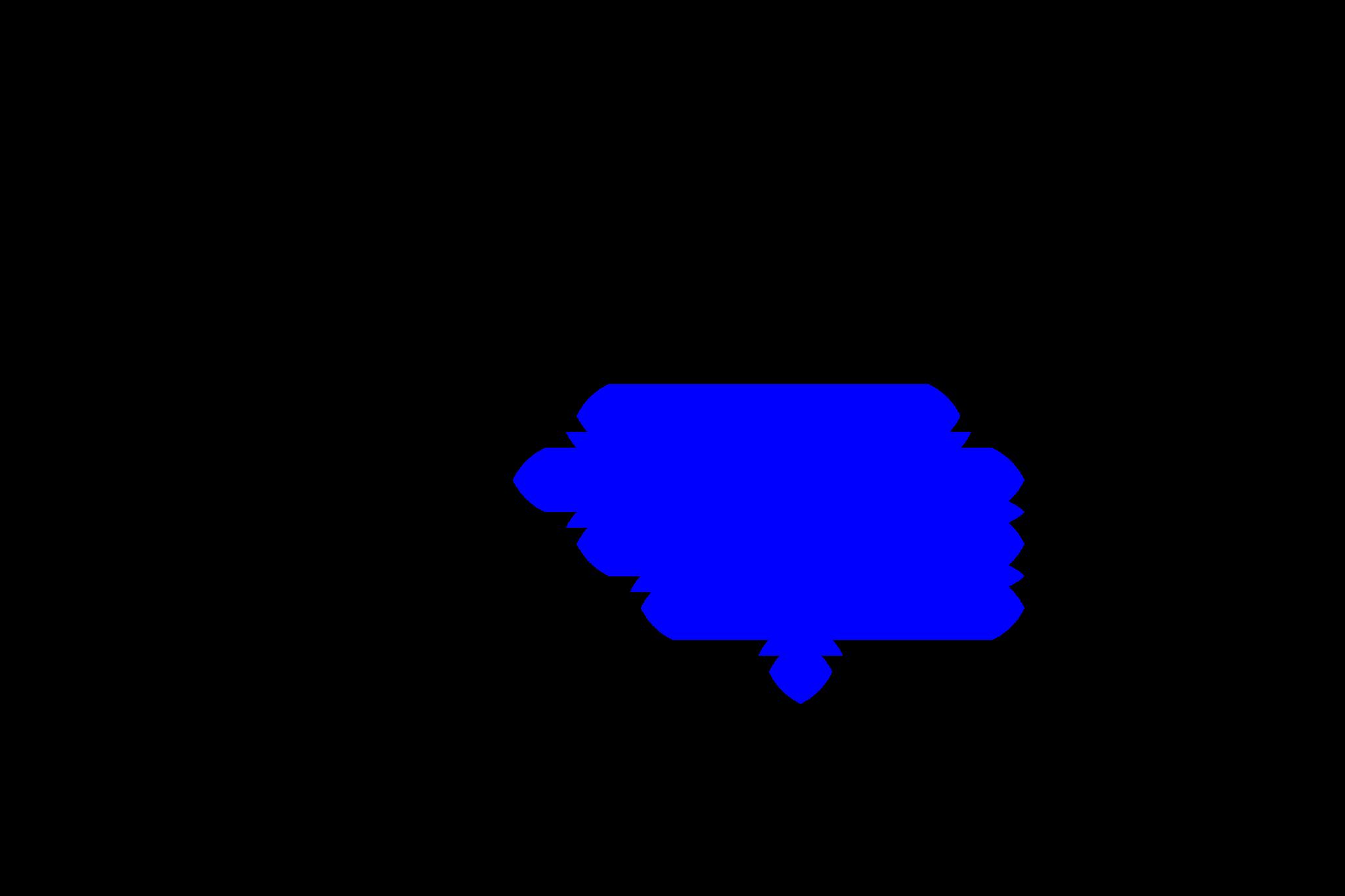}
  \end{subfigure}
  \begin{subfigure}[b]{\qualiwidth\linewidth}
  \includegraphics[width=\linewidth]{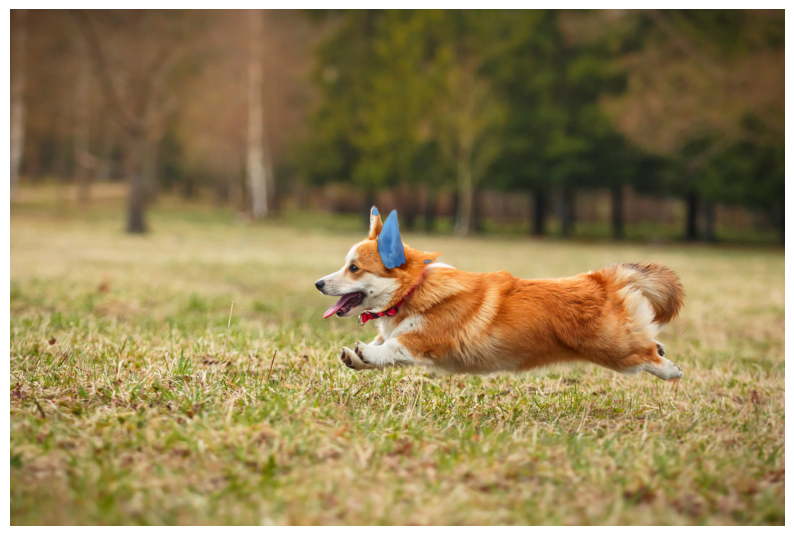}
  \end{subfigure}
  \begin{subfigure}[b]{\qualiwidth\linewidth}
 \includegraphics[width=\linewidth]{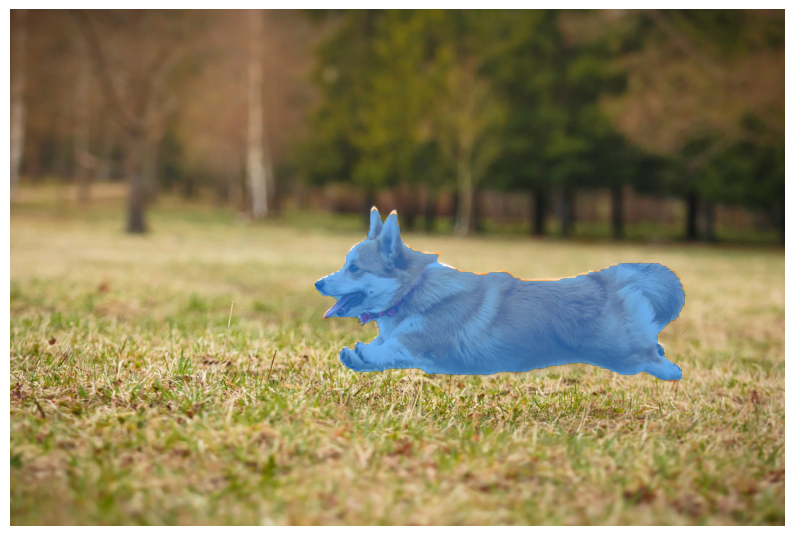}
  \end{subfigure}
  \begin{subfigure}[b]{\qualiwidth\linewidth}
  \includegraphics[width=\linewidth]{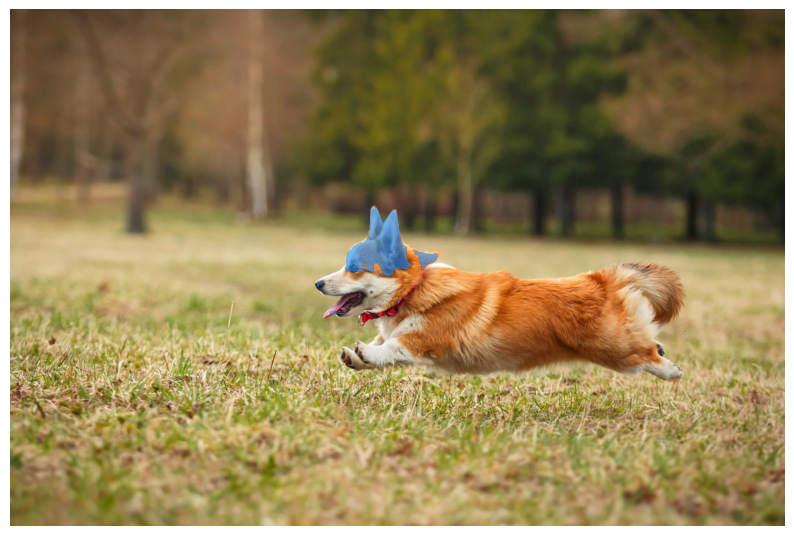}
  \end{subfigure}
  \begin{subfigure}[b]{\qualiwidth\linewidth}
 \includegraphics[width=\linewidth]{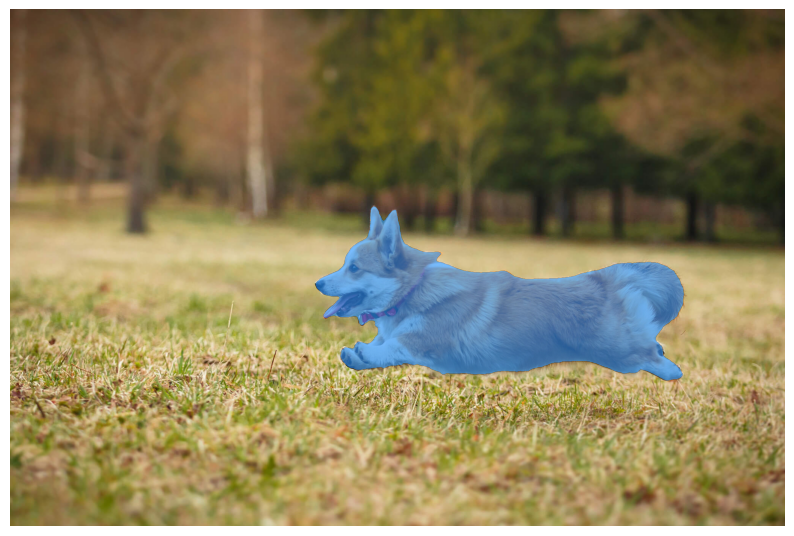}
  \end{subfigure}

\begin{subfigure}[b]{\qualiwidth\linewidth}
  \includegraphics[width=\linewidth]{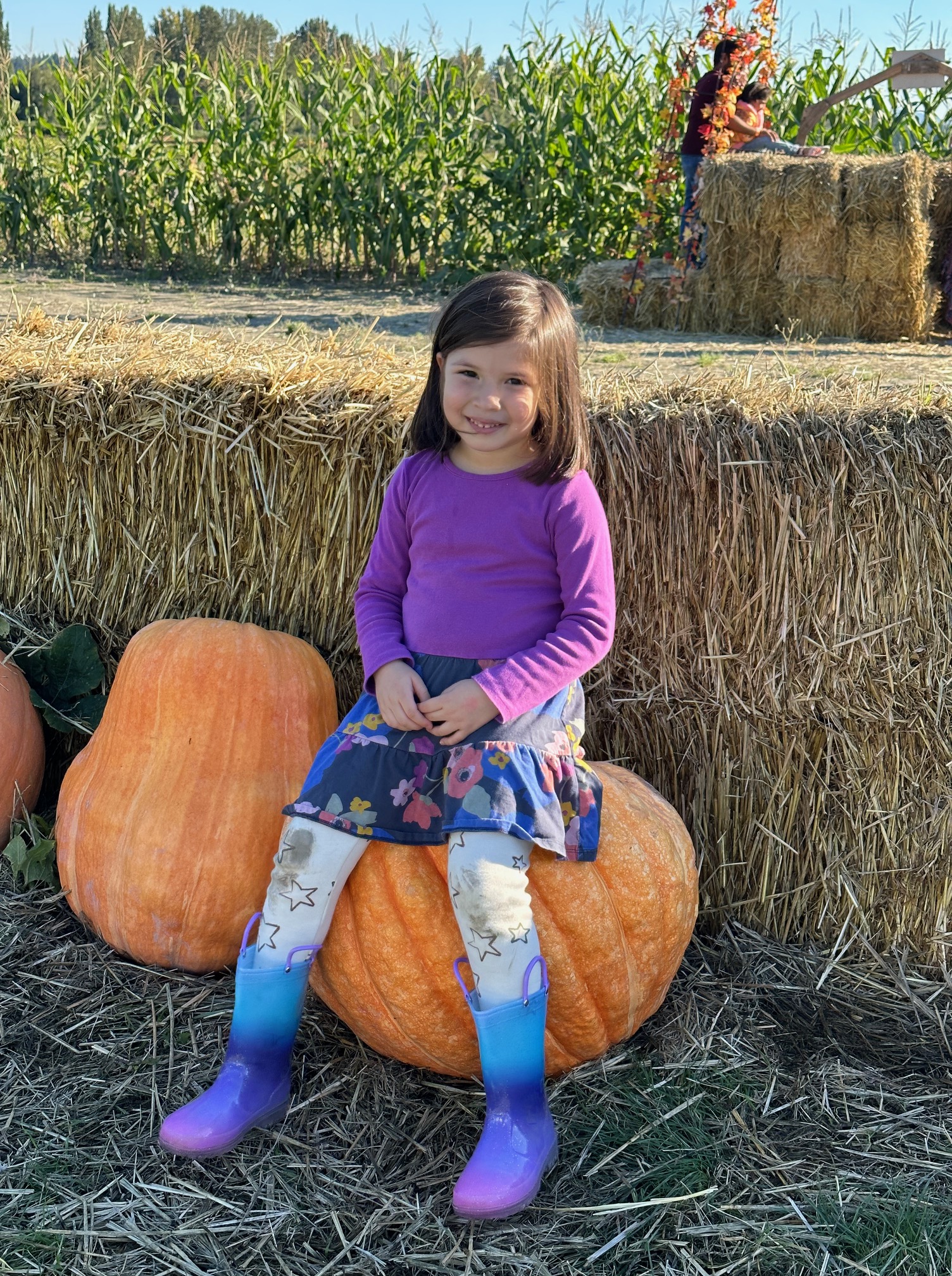}
  \end{subfigure}
\begin{subfigure}[b]{\qualiwidth\linewidth}
 \includegraphics[width=\linewidth]{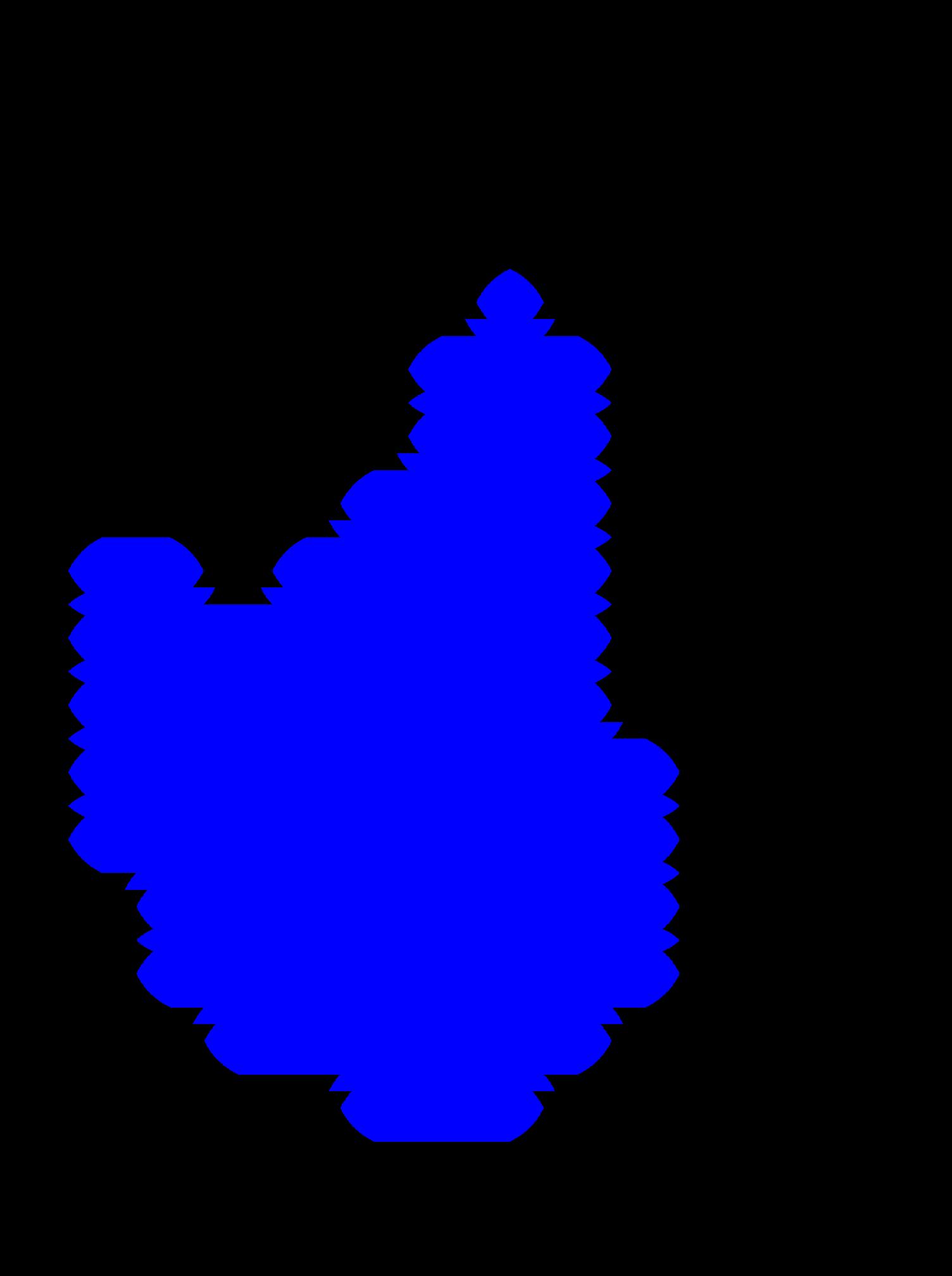}
  \end{subfigure}
  \begin{subfigure}[b]{\qualiwidth\linewidth}
  \includegraphics[width=\linewidth]{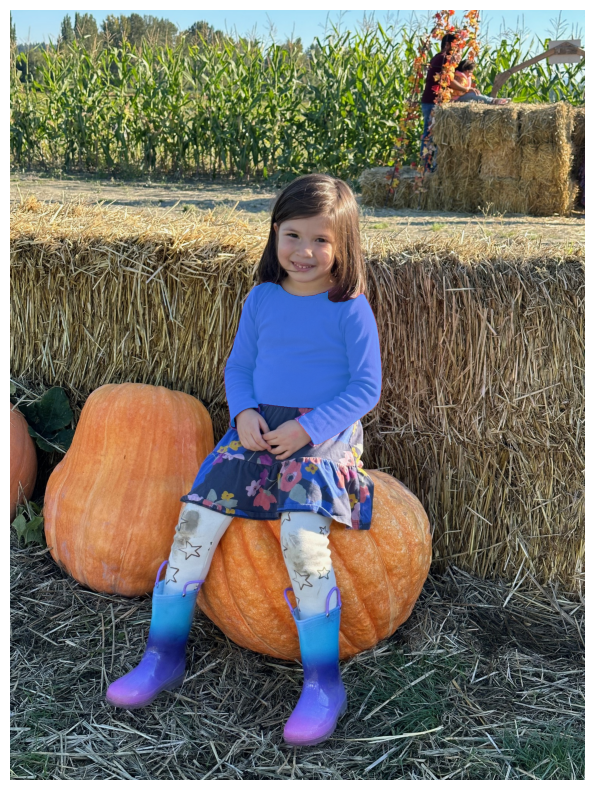}
  \end{subfigure}
  \begin{subfigure}[b]{\qualiwidth\linewidth}
 \includegraphics[width=\linewidth]{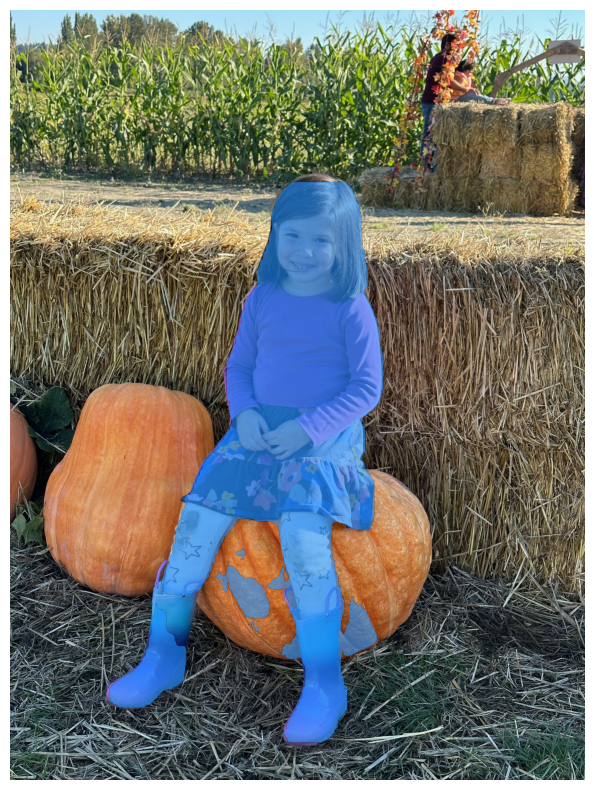}
  \end{subfigure}
  \begin{subfigure}[b]{\qualiwidth\linewidth}
  \includegraphics[width=\linewidth]{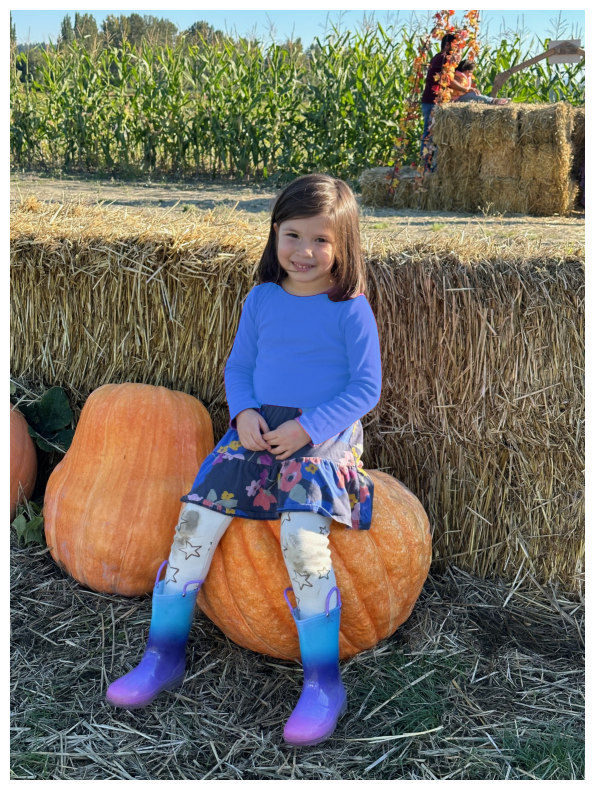}
  \end{subfigure}
  \begin{subfigure}[b]{\qualiwidth\linewidth}
 \includegraphics[width=\linewidth]{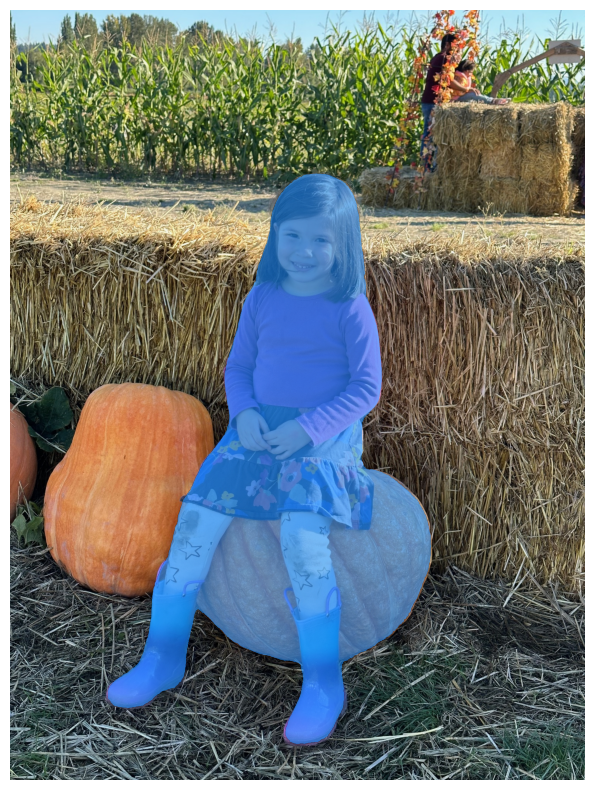}
  \end{subfigure}

\begin{subfigure}[b]{\qualiwidth\linewidth}
  \includegraphics[width=\linewidth]{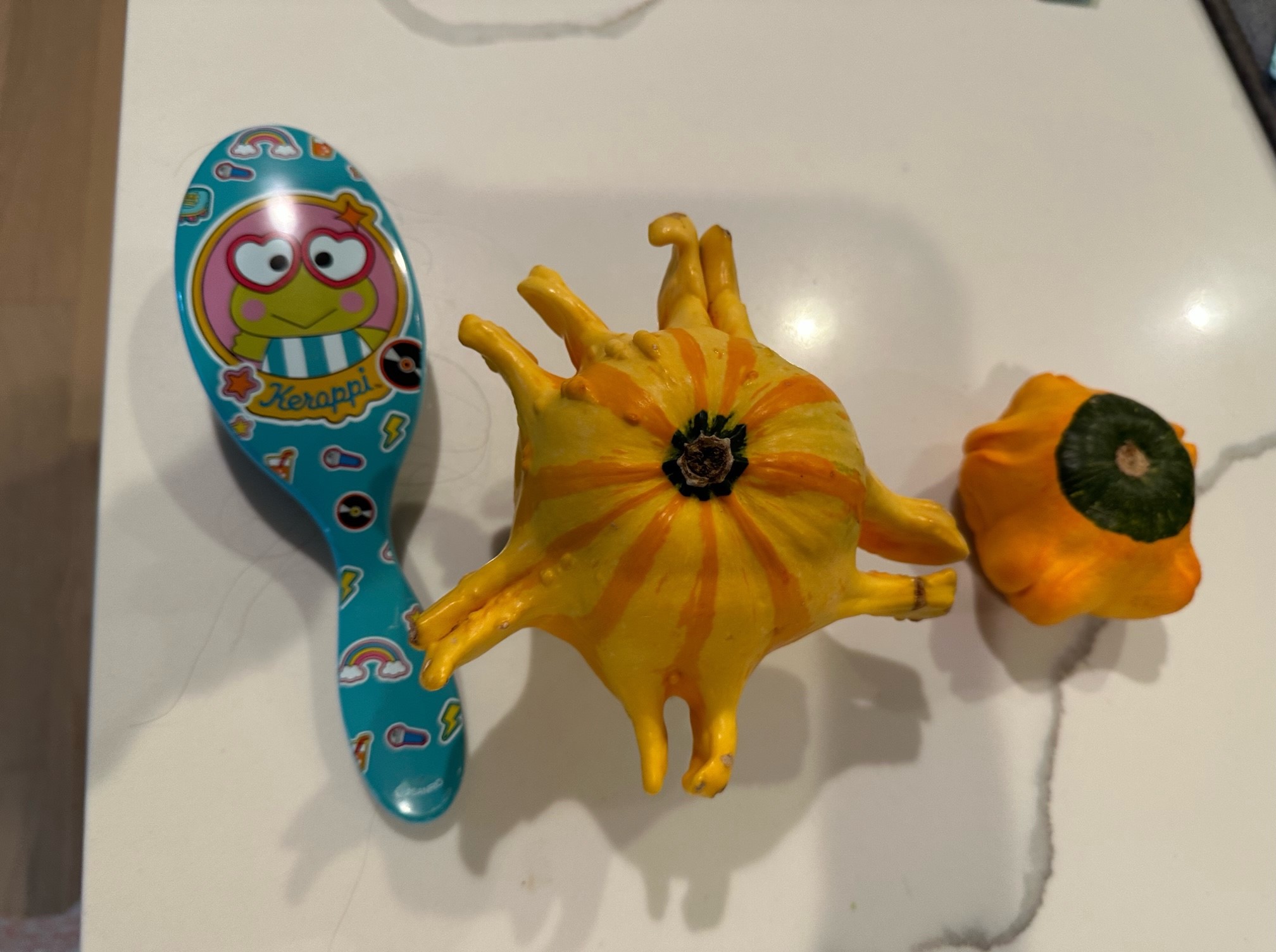}
  \end{subfigure}
\begin{subfigure}[b]{\qualiwidth\linewidth}
 \includegraphics[width=\linewidth]{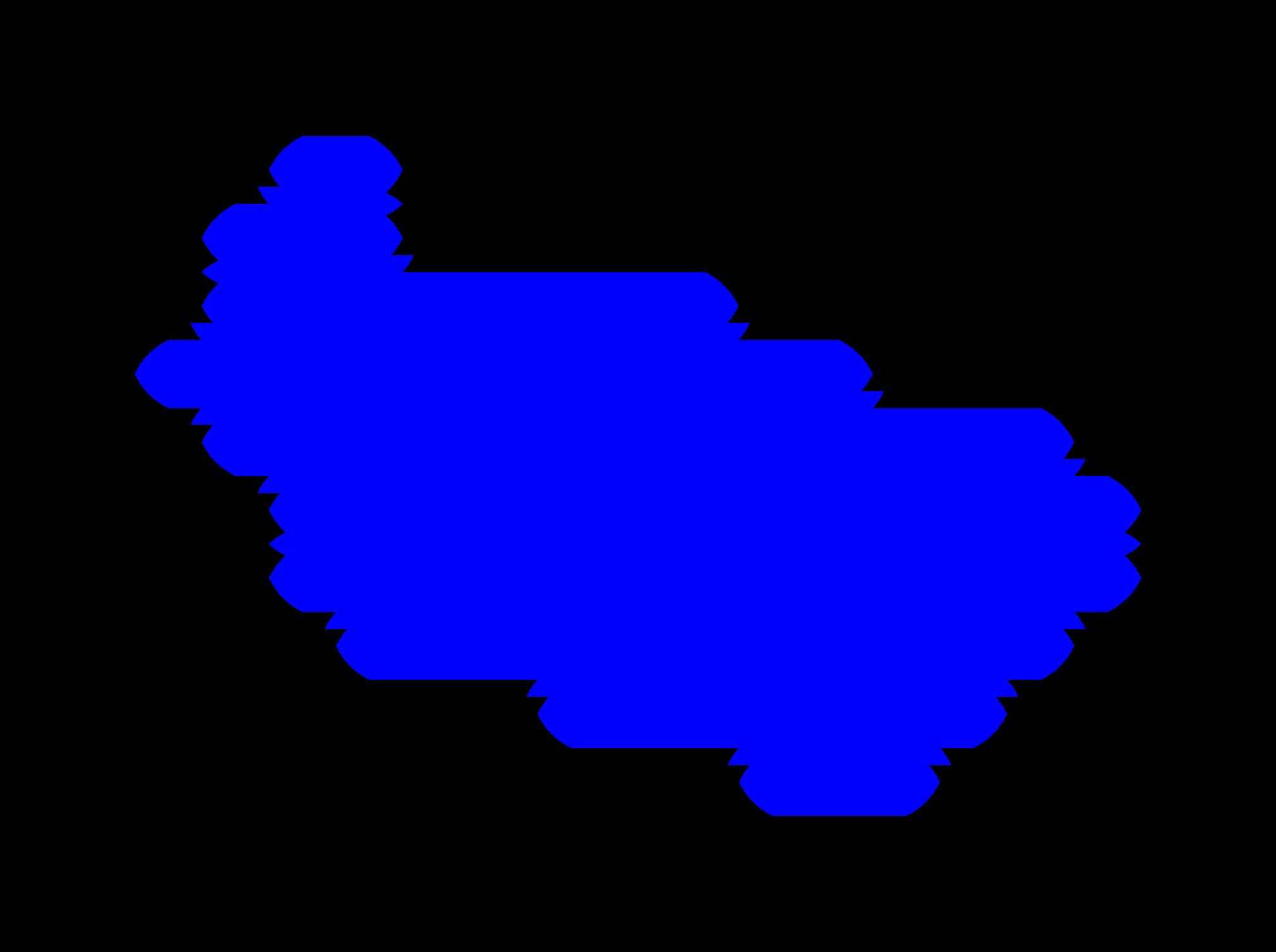}
  \end{subfigure}
  \begin{subfigure}[b]{\qualiwidth\linewidth}
  \includegraphics[width=\linewidth]{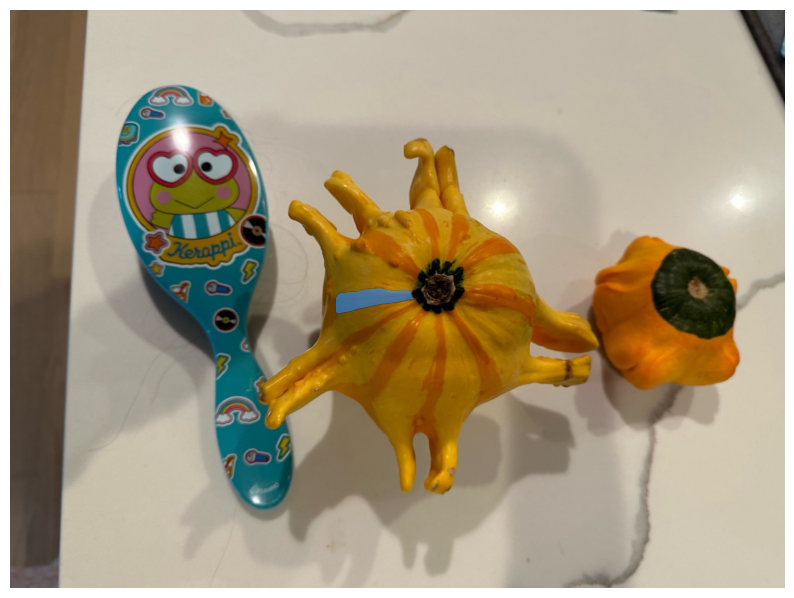}
  \end{subfigure}
  \begin{subfigure}[b]{\qualiwidth\linewidth}
 \includegraphics[width=\linewidth]{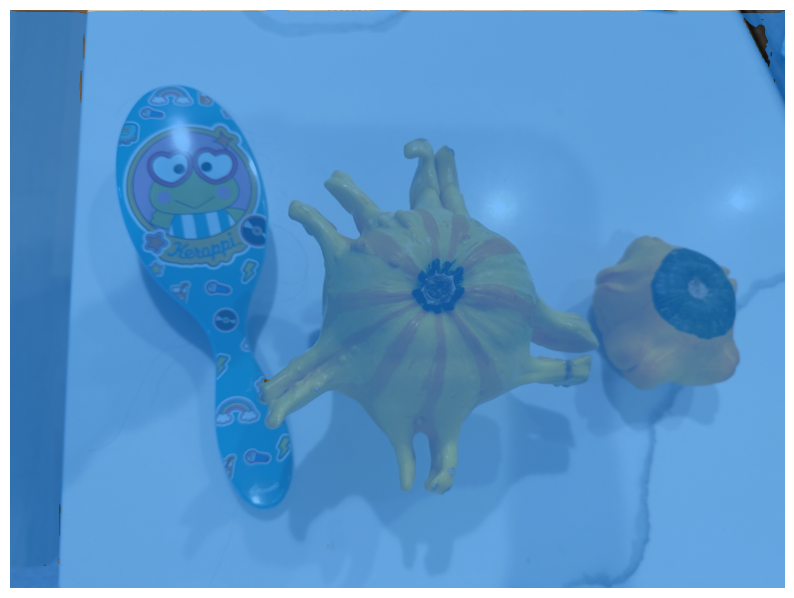}
  \end{subfigure}
  \begin{subfigure}[b]{\qualiwidth\linewidth}
  \includegraphics[width=\linewidth]{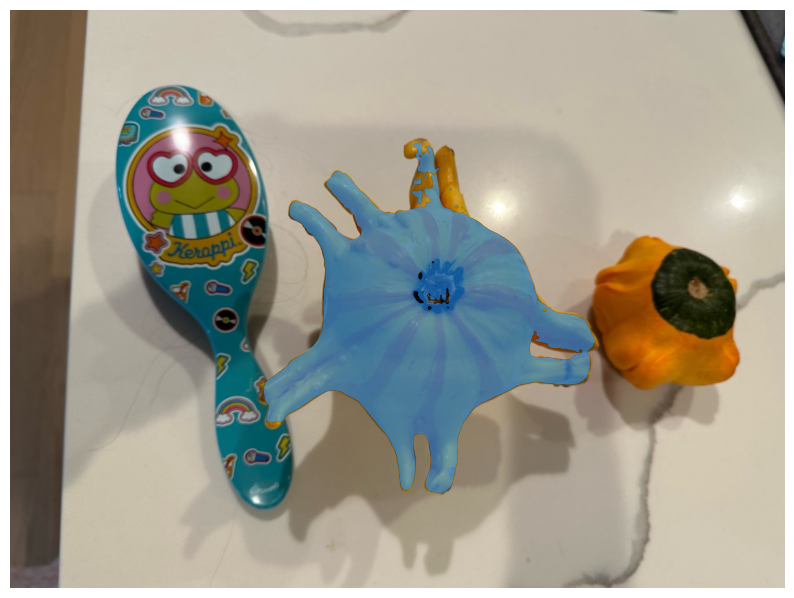}
  \end{subfigure}
  \begin{subfigure}[b]{\qualiwidth\linewidth}
 \includegraphics[width=\linewidth]{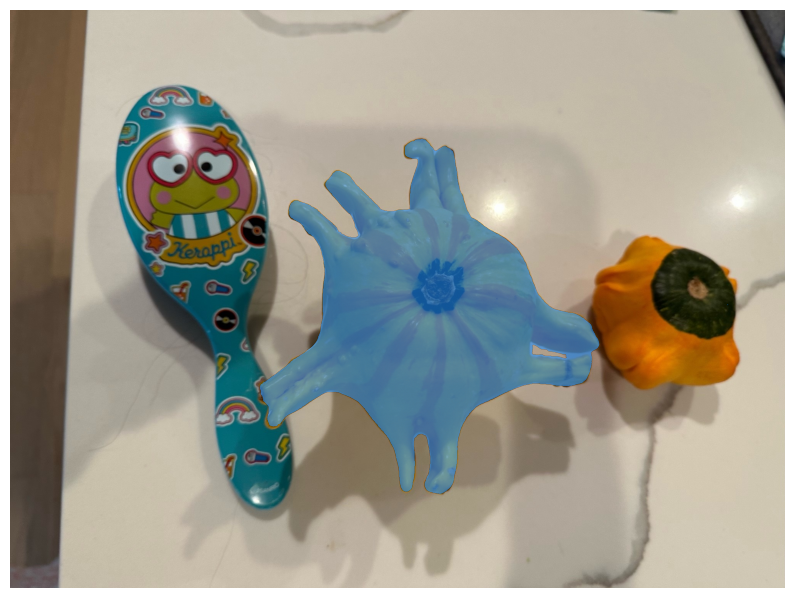}
  \end{subfigure}

\begin{subfigure}[b]{\qualiwidth\linewidth}
  \includegraphics[width=\linewidth]{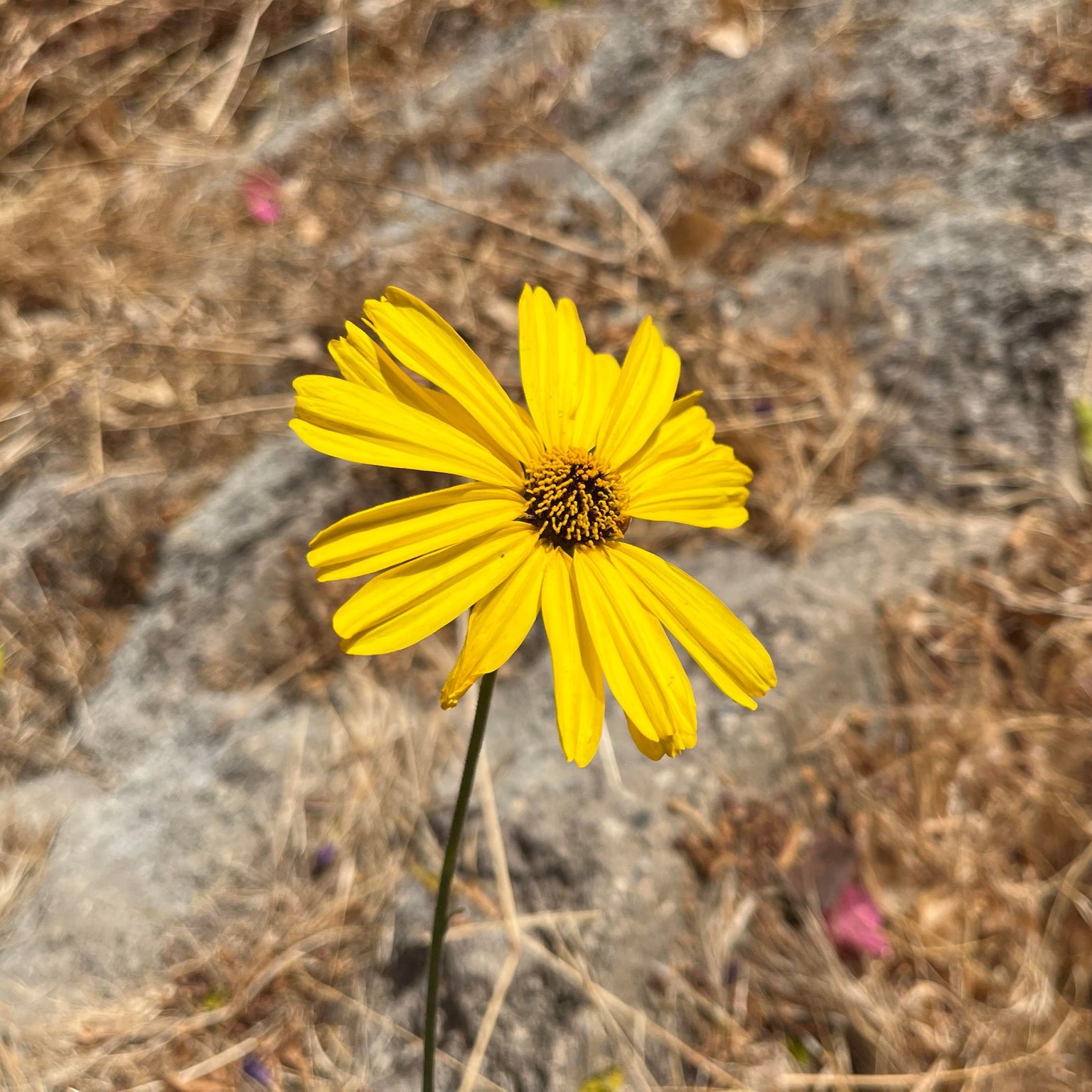}
  \subcaption{Input}
  \end{subfigure}
\begin{subfigure}[b]{\qualiwidth\linewidth}
 \includegraphics[width=\linewidth]{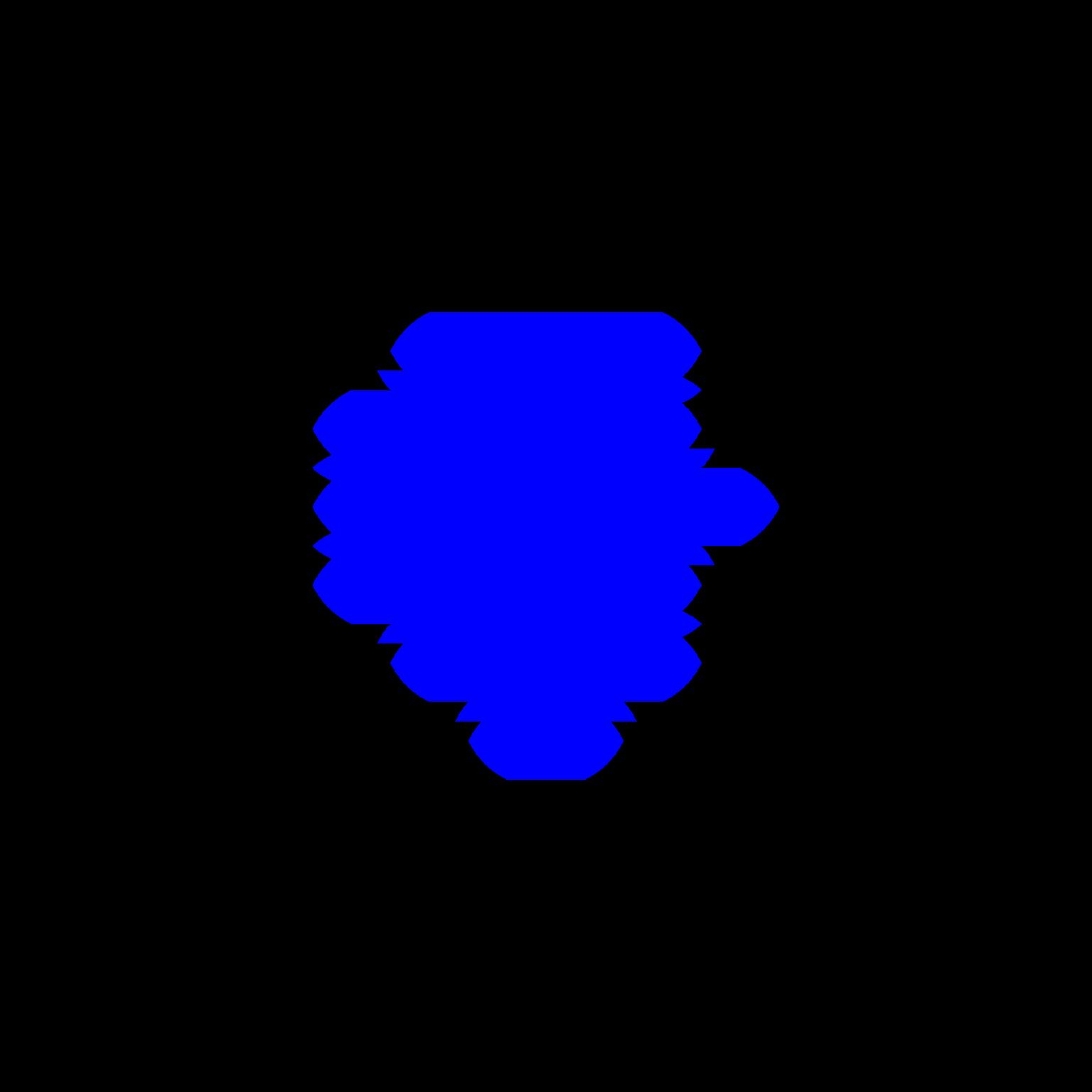}
 \subcaption{Saliency}
  \end{subfigure}
  \begin{subfigure}[b]{\qualiwidth\linewidth}
 \includegraphics[width=\linewidth]{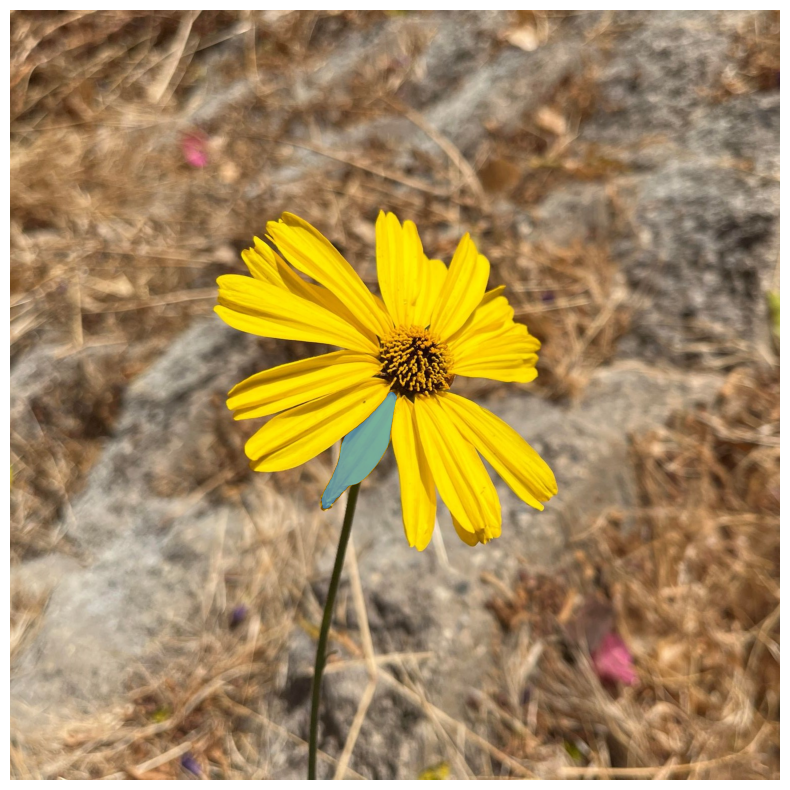}
  \subcaption{SAM~\cite{Kirillov2023_SAM}}
  \end{subfigure}
  \begin{subfigure}[b]{\qualiwidth\linewidth}
 \includegraphics[width=\linewidth]{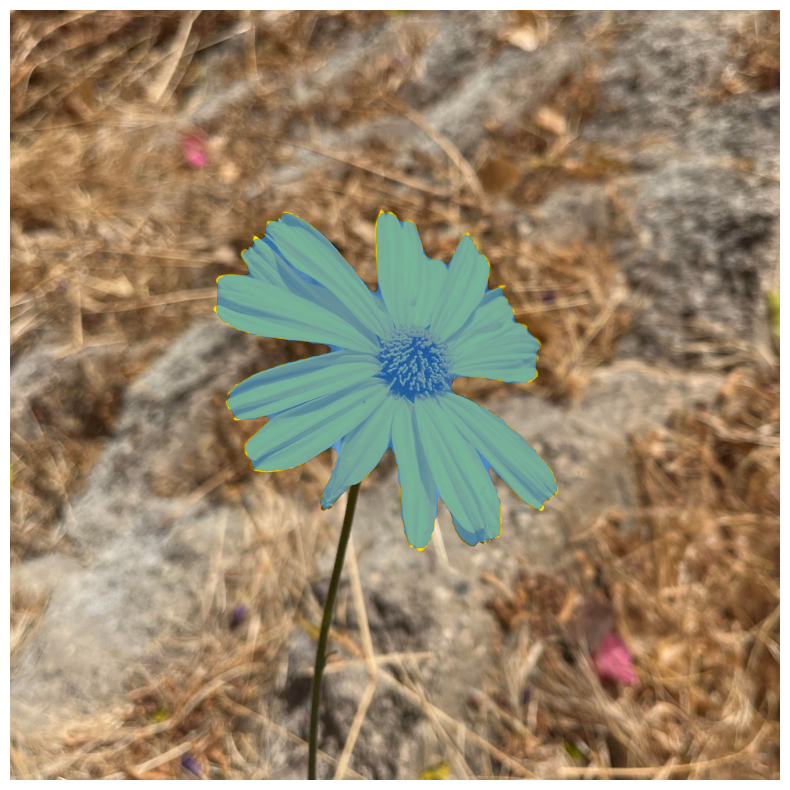}
 \subcaption{FastSAM~\cite{zhao2023_FastSAM}}
  \end{subfigure}
  \begin{subfigure}[b]{\qualiwidth\linewidth}
  \includegraphics[width=\linewidth]{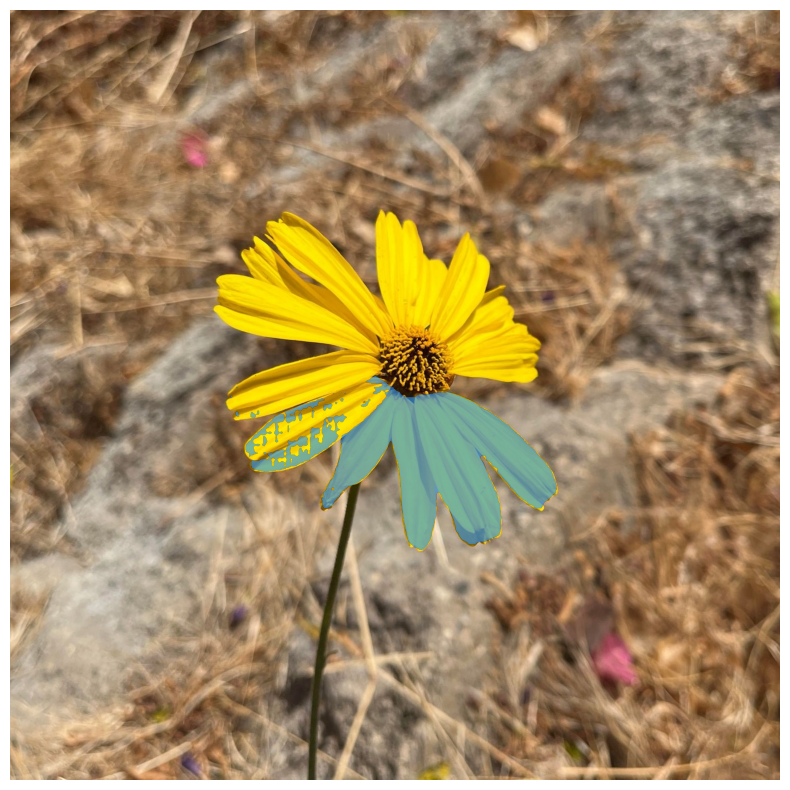}
  \subcaption{MobileSAM~\cite{zhang2023_MobileSAM}}
  \end{subfigure}
  \begin{subfigure}[b]{\qualiwidth\linewidth}
 \includegraphics[width=\linewidth]{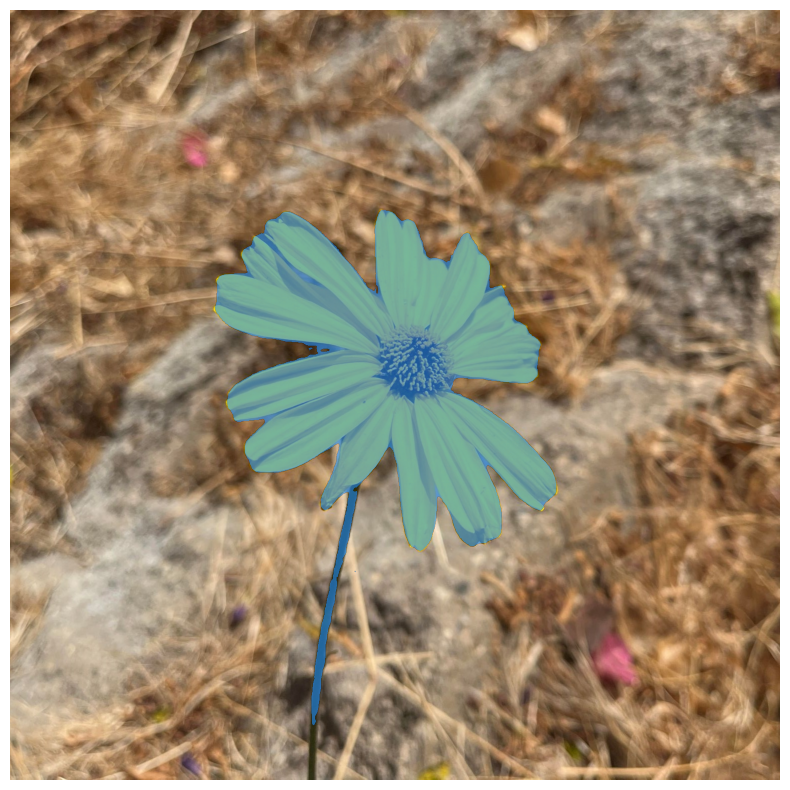}
 \subcaption{SqueezeSAM (Ours)}
  \end{subfigure}
\end{center}

\caption{Qualitative comparison with SOTA methods. From left to right: (1) Input image (2) Salient blob (3-5) Other model outputs (6) Our model. Note that, while all models missed cutting out salient objects simultaneously in row 3 (pumpkin and the kid), and missed the stem and the parts of the flower in row 5, SqueezeSAM successfully cut out related parts together.}
\label{fig:vsi_comp}
% \vspace{-3mm}
\end{figure*}

\section{Application}
\label{sec:salient_squeezesam}
The automatic generation of segmentation plays a crucial role in the development process for photography applications, as evidenced by its adoption by leading industry players such as Apple, CapCut. In line with these objectives, we explored 'classic' (non-interactive) instance segmentation and salient object detection (SOD) to facilitate auto-creation. Our qualitative observations indicated that SOD helped us create superior input points, leading to an enhanced interactive segmentation experience. An additional advantage of SOD is its ability to identify objects beyond the vocabulary of the training set, often including multiple contextually correlated objects, such as a person holding a parrot, rather than just a person or a common object. Consequently, we opted to utilize SOD in our approach.

\begin{figure}
    \centering
        \includegraphics[width=80mm]{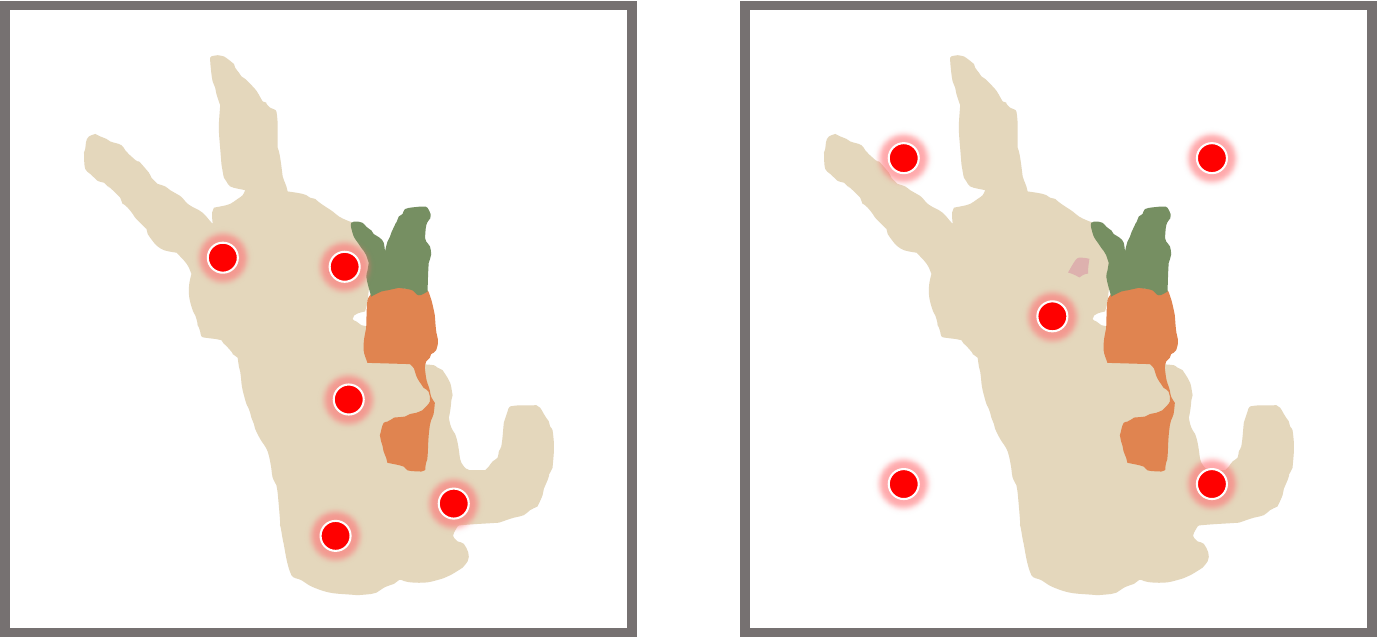}
    \caption{Left: Auto created clicks with respect to the center of mass of a blob; Right: Grid Sampling. Our proposed sampling method can provide better prompts to the interactive segmentation model than uniform sampling.}
    \label{fig:clicks}
    % \vspace{-2mm}
\end{figure}

The output of a SOD is a heatmap where the highest values of the heatmap refer to the most salient regions. To sample points from the saliency heatmap, after applying a global dynamic thresholding method like Otsu's method~\cite{2004_image_thresholding}, we detected the blobs. Most salient blob is identified with respect to the highest value within each blob, and we used the frequency of max heatmap value as a tie breaker when two blobs have the same value. After selecting the most salient blob, we divide it into 4 sections with respect the blob's center of mass. Then we use the center of mass from each section as our selected points besides the center of mass of the whole blob, creating 5 clicks. When the center of mass lies out of the heatmap, we use the closest point within the mask. This way of sampling guarantees that the point distribution is in accordance to the blob shape. Figure~\ref{fig:clicks} shows a comparison to the grid sampling.

\begin{figure*}[htbp]
\captionsetup[subfigure]{labelformat=empty}
\begin{center}
  \begin{subfigure}[b]{\qualiwidth\linewidth}
  \includegraphics[width=\linewidth]{imgs/comp/input/baseball_image.jpg}
  \end{subfigure}
\begin{subfigure}[b]{\qualiwidth\linewidth}
 \includegraphics[width=\linewidth]{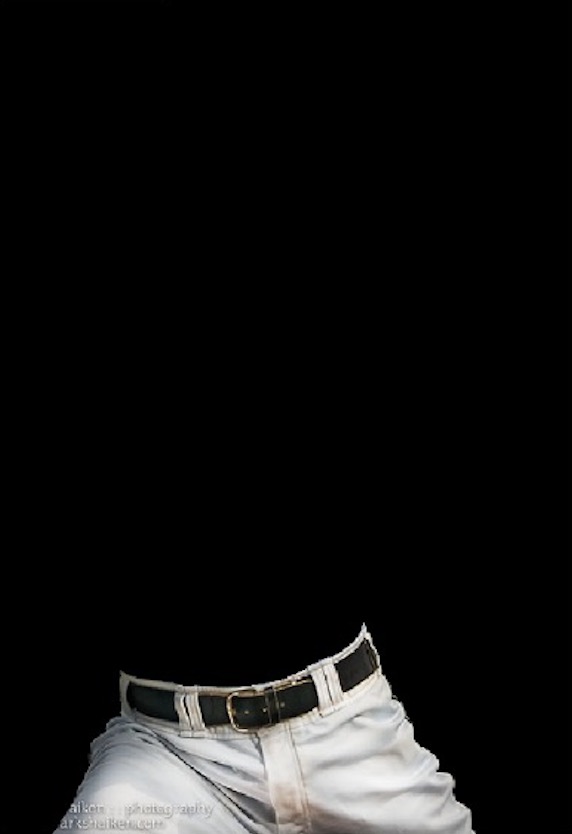}
  \end{subfigure}
  \begin{subfigure}[b]{\qualiwidth\linewidth}
  \includegraphics[width=\linewidth]{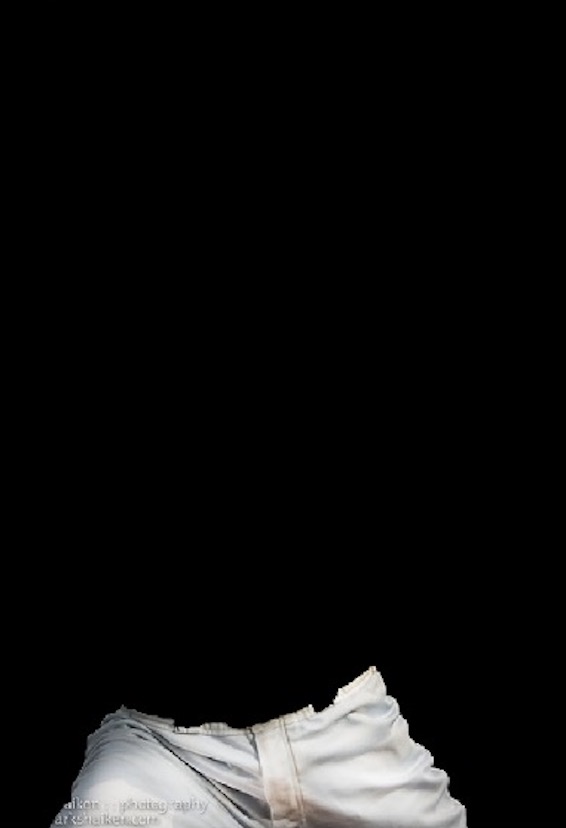}
  \end{subfigure}
  \begin{subfigure}[b]{\qualiwidth\linewidth}
 \includegraphics[width=\linewidth]{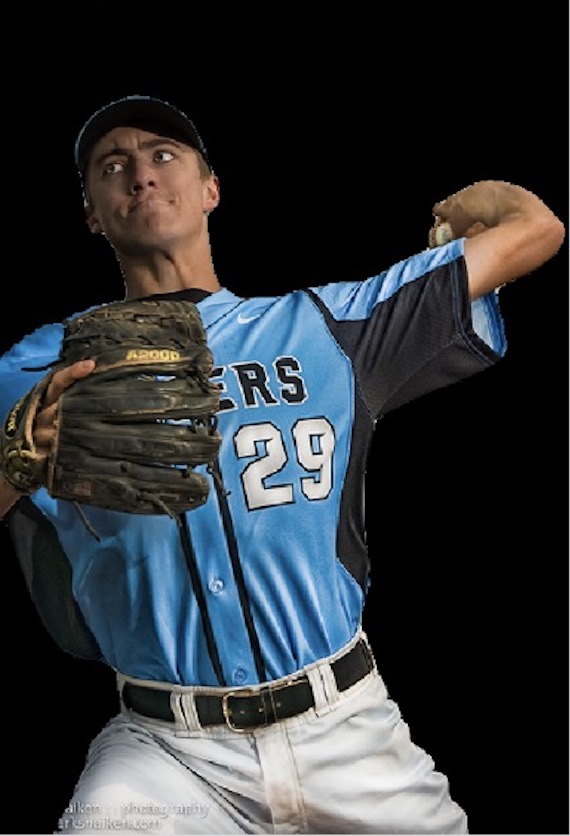}
  \end{subfigure}

  \begin{subfigure}[b]{\qualiwidth\linewidth}
  \includegraphics[width=\linewidth]{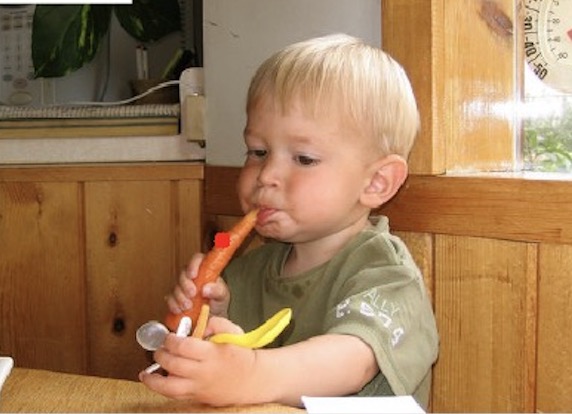}
  \end{subfigure}
\begin{subfigure}[b]{\qualiwidth\linewidth}
 \includegraphics[width=\linewidth]{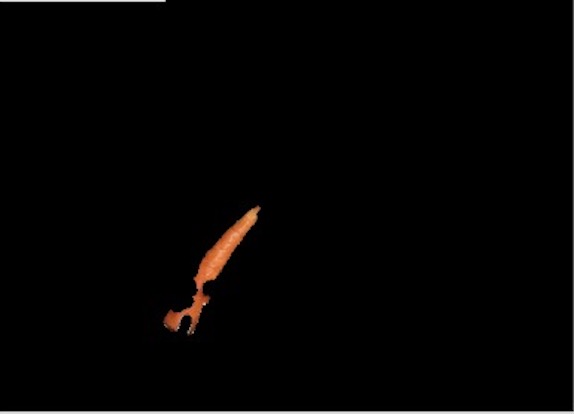}
  \end{subfigure}
  \begin{subfigure}[b]{\qualiwidth\linewidth}
  \includegraphics[width=\linewidth]{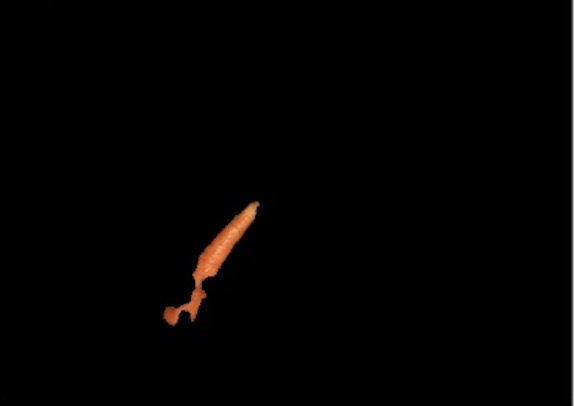}
  \end{subfigure}
  \begin{subfigure}[b]{\qualiwidth\linewidth}
 \includegraphics[width=\linewidth]{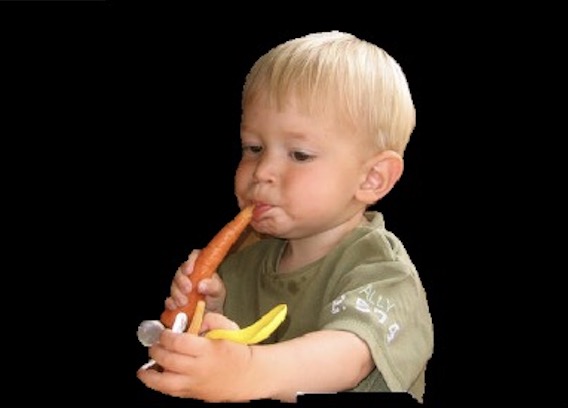}
  \end{subfigure}
\end{center}

\caption{From left to right: (1) User-click on original image (red-dot), (2) ground-truth mask from LVIS,  (3) original SAM output, (4) Salient-SqueezeSAM output. Notice that Salient-SqueezeSAM segments the all correlated objects, while other models only segment single object instances.}
\label{fig:vsi_comp_baseball}
% \vspace{-3mm}
\end{figure*}

% We have also observed that many users wish to segment people and pets when using mobile interactive segmentation.
For our application, we found that users are most interested in interactively segmenting people (and the things that people touch and hold) and pets (and dogs carrying frisbees and things like that).
We call this ``whole object segmentation."
However, with original SAM (or SqueezeSAM from the previous section), when you click on an object, it often segments just part of the object - see Figure~\ref{fig:vsi_comp_baseball}.
We address this with a series of data augmentations.
And we will show that these augmentations improve performance on whole object segmentation, with the tradeoff that they reduce performance when evaluated on the original COCO and LVIS evaluation sets.

%\begin{figure*}[htbp]
%\captionsetup[subfigure]{labelformat=empty}
%\begin{center}
%  \begin{subfigure}[b]{\qualiwidth\linewidth}
%  \includegraphics[width=\linewidth]{imgs/comp/input/child_carrot_image.jpg}
%  \end{subfigure}
%\begin{subfigure}[b]{\qualiwidth\linewidth}
% \includegraphics[width=\linewidth]{imgs/comp/input/child_carrot_gt.jpg}
%  \end{subfigure}
%  \begin{subfigure}[b]{\qualiwidth\linewidth}
%  \includegraphics[width=\linewidth]{imgs/comp/input/child_carrot_vith.jpg}
%  \end{subfigure}
%  \begin{subfigure}[b]{\qualiwidth\linewidth}
% \includegraphics[width=\linewidth]{imgs/comp/input/child_carrot_ours.jpg}
%  \end{subfigure}
%\end{center}
%
%\caption{From left to right: (1) A user clicks on a carrot, (2) ground-truth %mask, (3) original SAM output, (4) Salient SqueezeSAM output. Notice that %Salient SqueezeSAM segments the child holding the carrot, while other models %just segment the carrot.}
%\label{fig:vsi_comp_salient}
%% \vspace{-3mm}
%\end{figure*}

\subsection{Data Augmentation}

We identified some issues with the SA1B dataset that are especially detrimental to the task of whole-object segmentation on people and pets.
\begin{itemize}
    \item Ambiguity: A single click may correspond to multiple objects. Although SAM has a way to mitigate this by outputting multiple masks per inference call, this is typically not enough.
    \item No good representation of salient object masks. Each image in the SA-1B dataset has a collection of masks which are mostly non central to the theme of the image.
    \item Incomplete masks: Many times the important mask in the image (example person) would be missing.
\end{itemize}

\noindent
To address these issues, we use the following data augmentation techniques.
\begin{itemize}
    \item {\bf Mask merging.} We suppressed all masks that lie within a larger mask. For example, the models trained on the original dataset would segment out t-shirt when clicked on person's body. Thus, the training mask will not contain masks for tshirt if the same image has the mask for the person wearing the tshirt. As a result, the training data for the model is less ambiguous resulting in higher quality for people, pets and whole objects.
    \item {\bf Outlier injection.} We occasionally sample background points which do not correspond to any object during training. This makes the model more robust to outlier points that are occasionally sampled from the saliency mask (since there is no guarantee that the saliency mask is strictly within the ground truth).
    \item {\bf Center cropping around random objects.} The SA1B dataset contains many objects per image. However our usecase requires segmenting out images where there is one or two salient objects. In order to bridge the distribution gap, we randomly sample an object in the image and crop part of the image around that chosen object. We do the center cropping with 50 percent probability.
\end{itemize}

\subsection{Evaluation}

\begin{table}[]
\centering
\caption{Comparison of mIOU on COCO using 3 randomly sampled points from gt mask}
\label{T:eval_mIOU}
% \resizebox{\textwidth}{!}{
% \begin{tabular}{l|cccc|cccc}

\begin{tabular}{l|cccc}
\hline
 & \multicolumn{4}{c}{COCO} \\ %& \multicolumn{4}{c}{LVIS} \\
 & overall & large & medium & small \\ %& \multicolumn{1}{c}{overall} & \multicolumn{1}{c}{large} & \multicolumn{1}{c}{medium} & \multicolumn{1}{c}{small} \\
 \hline
Original SAM & 68.79\% & 76.12\% & 74.00\% & 60.52\% \\ \hline % &  &  &  &  \\ \hline
MobileSAM & 59.69\% & 73.86\% & 66.55\% & 45.92\% \\ % & 54.6\% & 76.56\% & 70.82\% & 40.73\% \\
SqueezeSAM & 62.62\% & 67.06\% & 63.65\% & 59.30\% \\ % & 62.57\% & 73.75\% & 72.34\% & 59.59\% \\
% Salient SqueezeSAM & 47.2\% & 73.1\% & 59.4\% & 39.1\% &  &  &  &  \\
% Salient SqueezeSAM (COCO/LVIS finetuned) & 54.22\% & 77.29\% & 60.04\% & 36.55\% & 41.43\% & 71.2\% & 58.8\% & 22.88\% \\
\hline
\end{tabular}

% }
\end{table}

Internally, we collected a dataset consisting of people images. We evaluate the quality of our model on this dataset in Table~\ref{tab:comp_sam}. We find that salient SAM significantly outperforms the standard SAM. This shows that tuning for saliency is helpful for improving quality of people.

\begin{table}
\centering
\caption{We collected an internal dataset of images containing person masks. These images mostly contain a single person as the central entity. We find that the salient SAM variants perform significantly better than the original SAM.}
\label{tab:comp_sam}
\begin{tabular}{c|c}
\hline
Model & mIOU \\
\hline
Original SAM & 80\% \\
\hline
Salient SqueezeSAM & 84\% \\
\hline
Salient SqueezeSAM (ft on LVIS) & 88\% \\
\hline
\end{tabular}
\end{table}

\section{Saliency Evaluation}
Each of our earlier evals focus on the nominal quality of the segmentation models. However, we would like to assess the quality of the segmentation models on salient objects. We use the following approach to determine whether an image has a salient object in each for each of the COCO images:
\begin{itemize}
    \item \emph{Saliency detection}: Run saliency detection on the image,
Candidate masks for a given image: For evaluation, we use the 5k validation partition of COCO. For each image, we use the masks present in the COCO instance segmentation. Furthermore, we augment the LVIS masks and replace the corresponding COCO mask if the new LVIS mask strongly overlaps with the COCO mask ($> 80\%$). Since the LVIS masks are more fine-grained and accurate around the edges, mIOU computed over the LVIS masks gives a better indication of quality.
\item \emph{Mask selection}: Pick the ground truth mask (among the various COCO/LVIS masks available for the image) that has max IOU with the saliency mask.
\item \emph{Addressing ambiguity in the saliency model}: Sometimes the saliency mask may be multimodal and could capture multiple objects. This introduces some ambiguity in the eval since our models are learned to predict single objects only. For fair comparisons, we need to ensure that the saliency mask is attending to the specific object (chosen from step 3 above). In order to address this, we  consider the chosen ground truth mask as valid only if there is sufficient intersection between the ground truth mask and the saliency mask. Our criteria for sufficient intersection is if at least 4 of the 5 gravity points chosen from the saliency mask lies within the ground truth mask. Otherwise, we consider the chosen mask to be ambiguous and all examples where there is ambiguity between the chosen ground truth mask and the saliency mask are ignored in the mIOU computation.
\item Compute the mIOU over all the examples chosen from step 4 of the 5k examples
\end{itemize}

Table~\ref{tab:saliency} compares various SAM models when queried using points sampled from the saliency mask.
Salient SqueezeSAM dramatically outperforms the prior literature.

\begin{table}
\caption{Comparison of various SAM models when queried using points from the saliency mask. The test set is the large masks from COCO and LVIS validation set.}
\label{tab:saliency}
\centering
\begin{tabular}{ccc}
\hline
Model & 1 point & 3 points \\
\hline
Original SAM & 56\% & 75\% \\
% \hline
Mobile SAM & 58\% & 76\% \\
% \hline
% Squeeze SAM & 31\% & 71\% \\
% \hline
Salient SqueezeSAM & 74\% & 74\% \\
% \hline
\end{tabular}
% \vspace{-2mm}
\end{table}

% \begin{table*}[]
% \caption{Comparison of mIOU quality. Each model is prompted with 3 input points. The points are randomly chosen from the ground truth mask.}
% \label{T:eval_mIOU}
% \footnotesize
% \centering
% \begin{tabular}{l|cccc|cccc}
% \hline
%  & \multicolumn{4}{c|}{COCO} & \multicolumn{4}{c}{LVIS} \\
%  & overall & large & medium & small & \multicolumn{1}{c}{overall} & \multicolumn{1}{c}{large} & \multicolumn{1}{c}{medium} & \multicolumn{1}{c}{small} \\ \hline
% Original SAM & 68.79\% & 76.12\% & 74.00\% & 60.52\% &  &  &  &  \\ \hline
% MobileSAM & 59.69\% & 73.86\% & 66.55\% & 45.92\% & 54.6\% & 76.56\% & 70.82\% & 40.73\% \\
% SqueezeSAM & 62.62\% & 67.06\% & 63.65\% & 59.30\% & 62.57\% & 73.75\% & 72.34\% & 59.59\% \\
% Salient SqueezeSAM &  &  &  &  &  &  &  &  \\
% Salient SqueezeSAM (COCO/LVIS ft) & 54.22\% & 77.29\% & 60.04\% & 36.55\% & 41.43\% & 71.2\% & 58.8\% & 22.88\% \\
% \hline
% \end{tabular}
% \end{table*}

Higher mIOU numbers with 1 click implies that we get the right object with a single click (for example person). The salient versions can get much higher quality on salient objects with a single click (ie they are able to identify the salient object with a single click rather than requiring more clicks). As expected fine-tuning on COCO and LVIS yields the best quality due to cleaner labels and adapting to the right distribution.

\section{Conclusions}
In this paper, our aim is to address the problem of segmentation in resource constrained environments, such as mobile devices, especially in the context of photo editing, where users want to cutout all the interesting parts of the image simultaneously, and it is non-intuitive to select each piece separately. For example, the expected cutout from an image of depicting a person riding a bike is usually the person together with the bike, or one may want to cutout a portrait image together with the hat on their head. Unfortunately, such an output cannot be obtained from neither semantic segmentation nor instance segmentation models. Segment Anything Model is designed to be category agnostic to segment any type of object, however it focuses on individual instances, rather than the semantic saliency. Our contributions can be summarized in three folds. 1) We propose the SqueezeSAM model, which performs drastically faster while being a tiny fraction of the size of the SAM. 2) We suggest to initialize the segmentation without any user input by utilizing a SOD model, whose output is processed to simulate the expected user-input to the segmentation model. 3) Our model can cutout correlated objects together to better capture the overall context. Because of its tiny size and tremendous speed, our Squeeze SAM can be deployed to any mobile device without throttling device resources, making it suitable for the common photo editing applications, easing the editing pipeline by auto-creating initial segmentation without requiring any user input.

% Please Do Not include Acknowledgements unless Camera Ready
% \section*{Acknowledgements}
% We would like to thank Eric Mintun and Nikhila Ravi for providing valuable feedback during the model iteration. We would like to thank Bochuan Du, Terry Know, Chirag Ramanlal,  Yang Qin and Avni Kakkar from the instagram team for providing valuable feedback on model quality when applied to instagram images. Finally would like to thank Lemeng Wu in help with creating visualizations and demo images.

% {\small
% \bibliographystyle{ieeetran}
% \bibliography{main}
% }

\end{document}